\RequirePackage{letltxmacro}
\documentclass[pdflatex,sn-mathphys]{sn-jnl}

\jyear{2024}%
\usepackage[utf8]{inputenc}
\usepackage[T5]{fontenc}
\usepackage{amsmath}
\usepackage{tabulary}
\usepackage{latexsym}
\usepackage{times}

\usepackage{float}
\usepackage{makecell}
\usepackage{array,booktabs,arydshln,xcolor}
\usepackage{subcaption,booktabs}
\usepackage{microtype}
\usepackage{multirow}
\usepackage{multicol}
\usepackage{threeparttable} 
\usepackage{amsmath}
\usepackage{tablefootnote}
\usepackage{pifont}
\usepackage{caption}
\usepackage{placeins}
\usepackage{array}
\usepackage{textcomp}
\usepackage{lmodern}
\usepackage{graphicx,graphics}
\usepackage{xcolor}
\usepackage{xurl}
\usepackage{amssymb}
\usepackage{pifont}
\usepackage{caption}
\usepackage{algorithm}
\usepackage{algpseudocode}
\usepackage{lipsum}

\definecolor{lavender}{RGB}{240,230,250}
\definecolor{lightblue}{RGB}{173, 216, 230}

\definecolor{britishracinggreen}{rgb}{0.0, 0.26, 0.15}
\definecolor{ao(english)}{rgb}{0.0, 0.5, 0.0}
\definecolor{applegreen}{rgb}{0.55, 0.71, 0.0}
\definecolor{lightred}{RGB}{255, 200, 200}

\theoremstyle{thmstyleone}%
\theoremstyle{thmstyletwo}%

\theoremstyle{thmstylethree}%

\begin{document}

\title[R2GQA System for Legal Regulations in Higher Education]{R2GQA: Retriever-Reader-Generator Question Answering System to Support Students Understanding Legal Regulations in Higher Education}



\author[1,2]{\fnm{Phuc-Tinh} \sur{Pham Do}}\email{20522020@gm.uit.edu.vn}

\author[1,2]{\fnm{Duy-Ngoc} \sur{Dinh Cao}}\email{20521661@gm.uit.edu.vn}

\author[1,2]{\fnm{Khanh} \sur{Quoc Tran}}\email{khanhtq@uit.edu.vn}

\author*[1,2]{\fnm{Kiet} \sur{Van Nguyen}}\email{kietnv@uit.edu.vn}

\affil[1]{\orgname{University of Information Technology}, \orgaddress{ \city{Ho Chi Minh City}, \country{Vietnam}}}

\affil[2]{\orgname{Vietnam National University}, \orgaddress{ \city{Ho Chi Minh City}, \country{Vietnam}}}

\abstract{In this article, we propose the R2GQA system, a Retriever-Reader-Generator Question Answering system, consisting of three main components: Document Retriever, Machine Reader, and Answer Generator. The Retriever module employs advanced information retrieval techniques to extract the context of articles from a dataset of legal regulation documents. The Machine Reader module utilizes state-of-the-art natural language understanding algorithms to comprehend the retrieved documents and extract answers. Finally, the Generator module synthesizes the extracted answers into concise and informative responses to questions of students regarding legal regulations. Furthermore, we built the ViRHE4QA dataset in the domain of university training regulations, comprising 9,758 question-answer pairs with a rigorous construction process. This is the first Vietnamese dataset in the higher regulations domain with various types of answers, both extractive and abstractive. In addition, the R2GQA system is the first system to offer abstractive answers in Vietnamese. This paper discusses the design and implementation of each module within the R2GQA system on the ViRHE4QA dataset, highlighting their functionalities and interactions. Furthermore, we present experimental results demonstrating the effectiveness and utility of the proposed system in supporting the comprehension of students of legal regulations in higher education settings. In general, the R2GQA system and the ViRHE4QA dataset promise to contribute significantly to related research and help students navigate complex legal documents and regulations, empowering them to make informed decisions and adhere to institutional policies effectively. Our dataset is available\footnote{Link for accessing to the dataset.} for research purposes.
}

\keywords{Question Answering, Retriever-Reader-Generator, Transformer, Legal Regulation, Higher Education   
}

\maketitle

\section{Introduction}\label{Introduction}
The educational regulations of universities consist of documents regarding training regulations, provisions, and guidelines on current training programs that students must adhere to to complete their academic programs. However, a significant challenge lies in the potential length and complexity of these educational regulations, making it difficult to read and extract information. Searching for specific information from these documents can be time-consuming, posing difficulties for students and lecturers. Alternatively, students may search for the wrong document, leading to misinterpretations and consequential adverse effects on students.

\begin{figure}[!ht]
    \centering
    \includegraphics[width=0.75\textwidth]{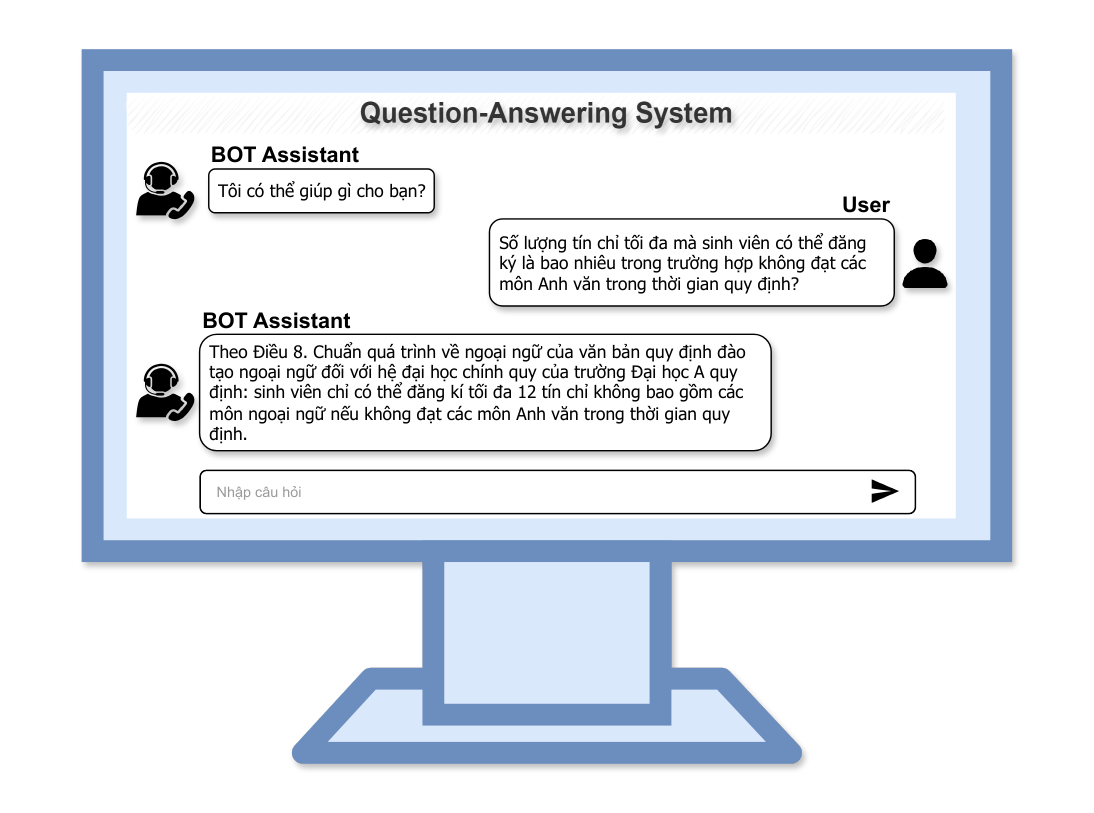}
    \caption{An example of the input and output of an educational regulations question-answering system. The input is the question of the user and the output is the response of the assistant. The response of assistant includes the answer to the question and the title of the document containing the answer.}
    \label{fig:example-task}
\end{figure}

The question-answering system (QAS) can help address the above problem. Similarly to search engines such as Google or Bing, the output of such a system is the answer to the input question based on the information available in the database. A question-answering system typically comprises two main components: a Document Retriever and a Machine Reader (or Answer Generator for abstractive questions). The document retriever queries relevant information related to the question, which is then passed to the Reader/Generator along with the question to generate an answer. Figure \ref{fig:example-task} shows the input of the USER question and the output of the BOT Assistant of the question answering system.

In the field of legal in Vietnamese, several question-answering systems have been developed. For example, in 2014, the vLawyer system was proposed by \cite{vLawyer}, a simple question-answering system with words in the answers directly extracted from documents. Therefore, it can be seen that very few question-answering systems in Vietnamese can provide answers with a human-like style.

When addressing abstract responses using language models, there are several approaches. An approach involves extracting multiple spans from the context and concatenating them as part of the MUSST framework \cite{Yang2020MultispanSE}. However, this method renders the responses less natural and diverse in language than human-like expressions. Another approach consists of passing the question and context through a generator module to produce complete answers (RAG). This method may result in less accurate output answers due to contextual overload, leading to noise. Furthermore, current answer generator models primarily perform text summarization tasks, which are not always suitable for answer extraction tasks. Enhancing performance can be achieved by using a machine reader module to extract answers before passing them through the answer generator models.

An essential component for implementing a question-answering system is training data. For the Vietnamese legal document, there are currently a few datasets available to build question answering systems. The dataset from Kien et al. (2020) \cite{kien-etal-2020-answering} and the dataset from Pham and Le (2023) \cite{from-4547} are two typical examples. However, the output of the two tasks is a set of ranked texts related to the question. Therefore, in Vietnamese, there is still a lack of datasets with answers extracted from various positions within the context or with natural language styles. Hence, constructing a training dataset with answers synthesized from multiple spans appearing in different positions within the context or having a natural, human-like style is very necessary. In this paper, we have three contributions:
\begin{itemize}
    \item \textbf{Question-answering system:} We designed a Retriever-Reader-Generator system named R2GQA, the first question-answering system for abstractive answers in Vietnamese, leveraging answers from the Machine Reader. Leveraging answers from the Reader and combining them with questions for the Generator to generate answers helps reduce noise compared to incorporating all information from the context.

    \item \textbf{Dataset construction:} We create a machine reading comprehension dataset named ViRHE4QA based on the legal regulations in higher education. This is the first Vietnamese dataset in the domain of university regulations, including various types of answers: multi-span extracted answers, abstractive answers. This dataset comprises 9,758 question-answer pairs that will be used to train the Reader and Generator models in our systems.

    \item \textbf{Experiments and evaluation:} We conduct experiments to evaluate models in the Document Retriever module. We also evaluate extractive reading comprehension models in the machine reader module and text generation models in the Generator module. Additionally, we analyze and compare the performance of our system with open-book question-answering systems, such as RAG \cite{RAG}.
\end{itemize}

The sections of the paper include Section \ref{Introduction} that provides an overview of the R2GQA system and the ViRHE4QA dataset. Section \ref{Related} reviews studies related to question-answering systems and datasets in the world and Vietnam. Section \ref{Corpus} describes the creation and characteristics of the ViRHE4QA dataset. Section \ref{QAS} details the design and implementation of our R2GQA system. Section \ref{Ex} presents the experimental setup and findings. Section \ref{discussion} interprets the results and highlights their implications. Section \ref{sec:error_analysis} examines the limitations and challenges encountered. Finally, Section \ref{sec:conclusion_future_work} summarizes the study and suggests directions for future research.

\section{Related Works}\label{Related}

\subsection{Related Question Answering System}
A question-answering system is a challenging task in natural language processing (NLP). Various types of systems have been developed to date. Based on the type of output answers, there are two popular systems in the field of NLP. 

First, question-answering systems with answers extracted from context (extractive question-answering). Some question-answering systems of this type include vLawyer \cite{vLawyer}, a simple question-answering system on Vietnamese legal texts proposed by Duong and Ho (2014) \cite{vLawyer}. vLawyer consists of two components: Question Processing and Answer Selection. 

DrQA, which is designed for reading comprehension in open-domain question-answering, as proposed by Chen et al. (2017) \cite{chen2017reading}. BERTserini \cite{bertserini} is a question-answering system that combines two models: BERT \cite{devlin-etal-2019-bert} and Anserini. Anserini is an information retrieval tool that identifies relevant documents that are likely to contain the answer. BERT \cite{devlin-etal-2019-bert} (Bidirectional Encoder Representations from Transformers) is a language model that understands context and the relationships between words to extract answers from context retrieved by Anserini. MUSST \cite{Yang2020MultispanSE} is a framework that is used to automatically extract answers from a given context. The answers of this framework are formed from multiple spans in the context to create human-like answers. This framework has two main modules: Passage Ranker and Question Answering. 

XLMRQA \cite{XLMRQA} is the first Vietnamese question-answering system with three modules: document retriever, machine reader, and answer selector). This question-answering system outperforms DrQA and BERTserini on the UIT-ViQuAD dataset \cite{ViQuAD}.  ViQAS is a question-answering system proposed by Nguyen et al. (2023) \cite{viqas}. In addition to the three retriever-reader-selector modules similar to XLMRQA, ViQAS includes an additional preprocessing rule step before the retriever module. Additionally, in the retriever module, the authors implemented smaller steps including evidence extraction and re-ranking. These changes contributed to ViQAS outperforming DrQA, BERTserini, and XLMRQA in the datasets UIT-ViQuAD \cite{ViQuAD}, ViNewsQA \cite{vinewsqa}, and ViWikiQA \cite{viwiki}. 

Second, question-answering systems with abstractive answer (abstractive question-answering). For this type of system, there are two common systems: open-book question answering and closed-book question answering.  Open-book question-answering systems typically have two modules: retriever and generator. The generator module in these systems is a sequence-to-sequence model such as T5 \cite{T5} or BART \cite{bart}. Some systems proposed based on open-book question answering include Fusion-in-Decoder \cite{FiD} and RAG \cite{RAG}. In the past two years, RAG has become very popular due to the strong development of large language models (LLMs) such as Gemini, GPT-4, or Copilot. These LLMs significantly enhance the performance of RAG due to their ability to generate accurate answers. 

Closed-book question-answering systems typically have one module, the generator. These generator models are usually generative language models like seq2seq pre-trained on a large collection of unsupervised texts. With enough parameters, these models can memorize some factual knowledge within their parameter weights. Therefore, these language models can freely generate answers to input questions without needing context. Some studies have used this method, such as the paper \cite{how-much}, and CGAP \cite{CGAP}. Recently, with the boom of LLMs such as GPT-3.5, GPT-4, Gemini, and LlaMa, closed-book question-answering has been widely applied in practice, and chatbots are increasingly appearing. However, the closed-book question-answering method can sometimes result in hallucination, causing confusion and inaccuracies in the answers.

Both of these methods have different advantages and disadvantages. Therefore, in this paper, we design a question-answering system with three modules (Retriever-Reader-Generator) to leverage the strengths and overcome the limitations of the aforementioned methods. This is the first question-answering system for abstractive answers in Vietnamese.
\subsection{Related Dataset}
Developing question-answering (QA) systems for specific domains requires specialized datasets tailored to domain knowledge and language. In the legal domain, several renowned datasets have been established and widely used. JEC-QA \cite{jecqa} is a comprehensive dataset comprising 26,365 multiple-choice questions, encompassing 13,341 single-answer questions (further divided into 4,603 knowledge-driven and 8,738 case-analysis questions) and 13,024 multi-answer questions (including 5,158 knowledge-driven and 7,866 case-analysis questions). BSARD \cite{BSARD}: BSARD was created by legal experts, BSARD comprises 1,108 questions derived from 22,633 legal articles. The dataset exhibits an average article length of 495 words, while questions range from 23 to 262 words, with a median length of 83 words. PRIVACYQA \cite{privacy}: PRIVACYQA comprises 1,750 questions spanning 335 policies and 4,947 sentences, meticulously crafted by experts. The dataset is characterized by its long texts, with an average document size of 3,237.37 words. In particular, PRIVACYQA encompasses a diverse range of question types, including unanswerable and subjective questions.

For Vietnamese, some work on legal QA datasets has recently been published. The QA data set was created by Kien et al. (2020) \cite{kien-etal-2020-answering} and includes 5,922 questions with 117,545 related articles. This is a large legal dataset for Vietnam. Each question has an average length of 12.5 words and is associated with 1.6 relevant articles. The dataset from Pham and Le (2020) \cite{from-4547} consists of 4,547 questions and 5,165 passages. The length of each pair of questions and answers is mostly less than 100 words. For the domain of university education regulations, the dataset from Phuc et al. (2023) \cite{IUH} comprises 10,000 data points in the training set and 1,600 in the test set, constructed based on the guidelines of the Ho Chi Minh City University of Industry. The answers in this dataset are extracted from contextual passages.

Therefore, currently there are few Vietnamese machine reading comprehension datasets containing answers that span multiple positions and exhibit human-like style in the domain of higher education regulations. We hope that our dataset will contribute additional resources for Vietnamese in creating and developing systems in this domain.
\section{Dataset}\label{Corpus}
\subsection{Dataset Creation}
\begin{figure}
    \centering
    \includegraphics[width=0.8\linewidth]{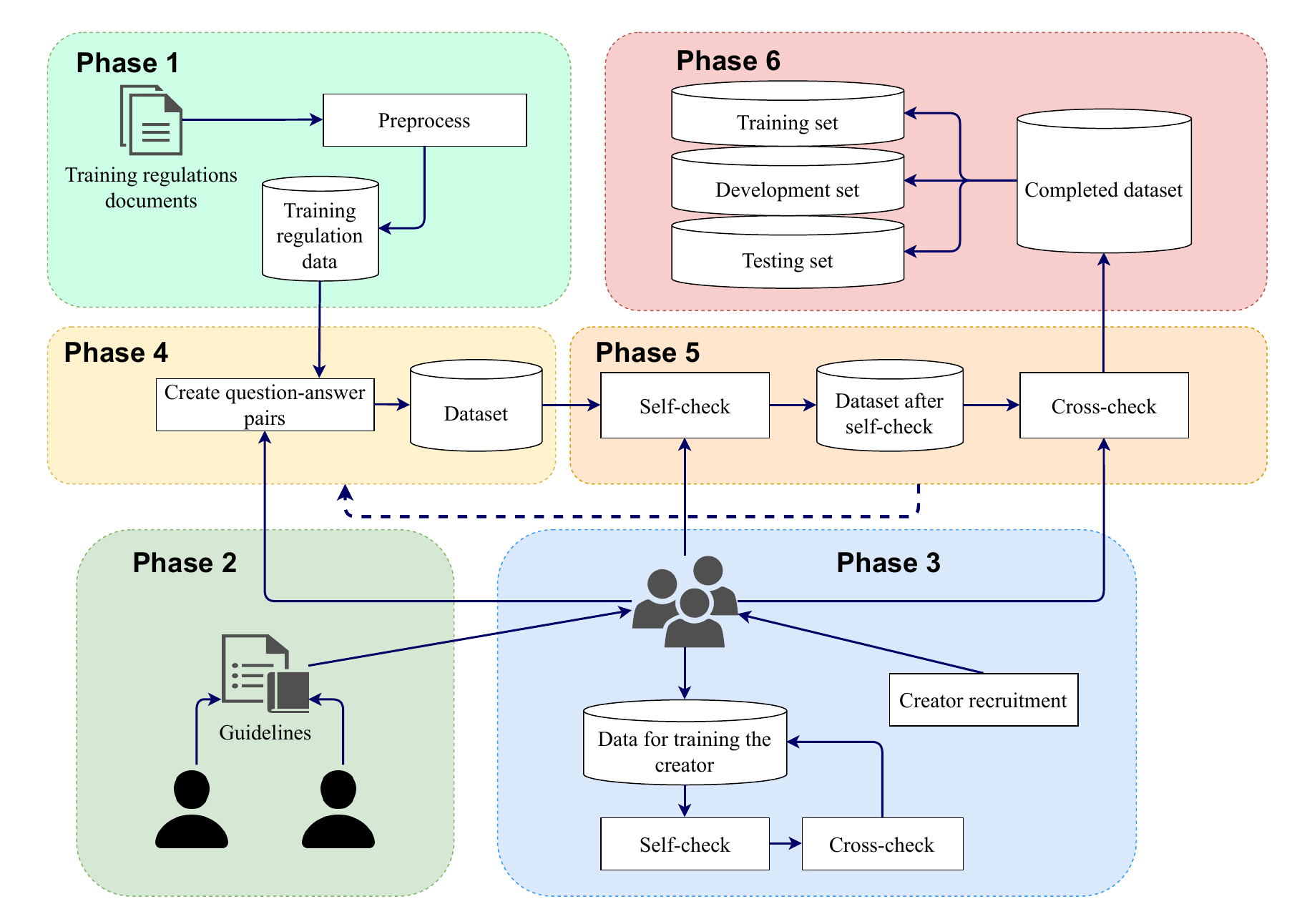}
    \caption{Dataset creation process. This process consists of 6 phases: context collection (Phase 1), guidelines creation (Phase 2), creator agreement (Phase 3), question-answer creation (Phase 4), data validation (Phase 5), and data splitting (Phase 6).}
    \label{fig:corpus-creation}
\end{figure}

In this section, we introduce how we constructed the dataset. Our dataset creation process consists of 6 phases: context collection (Section \ref{subsec:context_collection}), guidelines creation (Section \ref{subsec:guidelines_creation}), creator agreement (Section \ref{subsec:creator_agreement}), question-answer creation (Section \ref{subsec:question-answer_creation}), data validation (Section \ref{subsec:data_validation}), and data splitting (Section \ref{subsec:data_splitting}). These six phases are illustrated in Figure \ref{fig:corpus-creation}.

\subsubsection{Context Collection} \label{subsec:context_collection}
We collected regulatory documents regarding the curriculum of a university in Vietnam. The documents were gathered in various formats, such as Word, PDFs, or images, so we converted them to Word using smallpdf.com\footnote{\url{https://smallpdf.com/pdf-to-word}} or manually retyped them if the PDF file contains images. After converting all documents to the Docs format, we converted the tables into paragraphs using a predefined format. 

Following this process, we obtained 21 documents with an average length of 10.67 pages and 3,881 words per document. As the word count in each document is too large for language models, we divided the documents into smaller paragraphs (called articles) for convenience in dataset construction and model training. In the result, we obtained 294 articles (referred to as contexts below) with an average length of 234.49 words.

\subsubsection{Guidelines Creation} \label{subsec:guidelines_creation}

We relied on the guidelines from two datasets, UIT-ViQuAD \cite{ViQuAD} and ViRe4MRC \cite{vire4mrc}. These guidelines describe and provide detailed examples to help creators understand how to create questions and answers for the given problem consistently. The guidelines clearly outline different question types including "How", "What", "Which", "Where", "Why", "When", "Who", Yes/No, and other types such as "How long", "How many". Definitions and examples of each type of question are presented in the Table \ref{definition_question_type}.

\begin{table}[!ht]
\centering
\caption{Definitions and Examples of Different Types of Questions.}\label{definition_question_type}
\resizebox{1\linewidth}{!}{
\begin{tabular}{lp{6cm}p{6cm}}
\hline
\textbf{Question type} & \textbf{Definition} & \textbf{Example} \\
\hline
How & Questions of this type inquire about the method to do something. & Quy trình tổ chức thi hình thức vấn đáp, đồ án diễn ra như thế nào? \textit{(How is the process of organizing a question-and-answer or project-based exam conducted?)} \\
\hline
What & Questions of this type focus on definitions, objects, or events. & Hình thức phổ biến để lấy ý kiến sinh viên là gì? \textit{(What are common methods for gathering student opinions?)}\\
\hline
Which & These questions involve choices, where the answer selects one or more options presented within the question. & Các seminar được thực hiện bằng tiếng Anh hay tiếng Việt? \textit{(Which language are seminars conducted in, English or Vietnamese?)} \\
\hline
Where & Questions whose answers refer to an actual location or position. & Thành phần, nhiệm vụ, quyền hạn của Hội đồng phúc tra được qui định ở đâu? \textit{(Where are the composition, tasks, and powers of the Review Council defined?)} \\
\hline
Why & Questions seeking the reason or motive behind something. & Tại sao P.ĐTĐH lại cần phải tổng hợp các ý kiến và trình cho Hiệu trưởng? \textit{(Why does the Academic Department need to synthesize opinions and present them to the President?)}\\
\hline
When & Questions regarding time. & Khi nào thì sinh viên được Trường cấp email? \textit{(When are students provided with school email accounts?)} \\
\hline
Who & Questions identifying a person or group of people. & Ai là người quyết định thành lập Ban Điều hành Công tác giáo trình? \textit{(Who decides to establish the Curriculum Operations Board?)} \\
\hline
Yes/No & These questions typically end with the word "không" and have a yes or no answer option. The answer is evidence upon which a Yes or No choice can be based for the question. & Thành phần tham gia Tổ soạn thảo có thể bao gồm 6 thành viên không? \textit{(Can the Editorial Board consist of 6 members?)} \\
\hline
Others &  Questions whose answers are not in the above groups. The most frequently asked questions are how long, how many, or how much. & Thời gian làm bài thi tối thiểu là bao lâu? \textit{(How long is the minimum duration to take the exam?)} \\
\hline
\end{tabular}}
\label{tab:questions}
\end{table}
 
The guidelines cover various strategies for asking questions, such as asking questions from general to specific, asking questions in the order of the context before posing questions whose answers appear at multiple places, and posing "Wh" questions before "Yes/No" questions.

In this paper, we divide the answers into two categories: "Extraction answers" and "Abstract answers". The extractive answers have two types: single-span and multi-span. The extractive answers must contain complete information and be as concise as possible while present in the context (article). In the case of multi-span answers, the spans must be semantic equivalence and should not be concatenated from different parts of the context to form a complete sentence. In Table \ref{tab:example_extractive_answer}, the correct extractive answers should be "học kỳ chính" \textit{("regular semester")} and "học kỳ hè" \textit{("summer semester")} because these two phrases have semantic equivalence. The answer "Trường có" \textit{("The University has")}, "học kỳ chính" \textit{(regular semester)}, "và" \textit{("and")}, "học kỳ hè" \textit{("summer semester")} is not acceptable because this answer attempts to form a complete sentence, resulting in the extracted words lacking semantic equivalence. Abstractive answers are rewritten answers from the question and extractive answers that resemble how a human would answer, with additional words and meanings to smooth out the extractive answer without changing the meaning or adding new information. We encourage creators to be creative with their writing style for abstract answers. In the example at Table \ref{tab:example_extractive_answer}, the abstractive answer could be: "Trường có các loại học kỳ: học kỳ chính và học kỳ hè" \textit{("The university has semester types: regular semesters and summer semesters.")}. In this case, the abstractive answer does not include a counting number like "Trường có \textcolor{red}{hai} loại học kỳ: học kỳ chính và học kỳ hè" \textit{("The University has \textcolor{red}{two} types of semesters: regular semesters and summer semesters.")} as this adds information not present in the question or extractive answer.

\begin{table}[!ht]
    \centering
    \renewcommand{\arraystretch}{1.2} 
    \caption{An example of extractive answer and abstractive answer in the guidelines.}
    \begin{tabular}{p{1\linewidth}} 
        \hline 
        \textbf{\textit{Article:}}\\
        Học kỳ là thời gian để sinh viên hoàn thành một số học phần của chương trình đào tạo. Một \colorbox{lightred}{\textcolor{blue}{\textbf{học kỳ chính}}} có 15 tuần thực học \colorbox{lightred}{\textbf{và}} 2 đến 3 tuần dành cho đánh giá hoạt động đào tạo (thi cuối kỳ, thi giữa kỳ, kiểm tra,…). Một \colorbox{lightred}{\textcolor{blue}{\textbf{học kỳ hè}}} có tối thiểu 5 tuần thực học và 1 tuần thi. Căn cứ vào tình hình thực tế mỗi năm, kế hoạch giảng dạy của học kỳ có thể được điều chỉnh theo quyết định của Hiệu trưởng. 
        Một năm học có 2 học kỳ chính. Tùy theo điều kiện, \colorbox{lightred}{\textbf{Trường có}} thể tổ chức thêm học kỳ hè. Việc đăng ký học phần học kỳ hè được quy định tại Điều 14 của quy chế này.\\
        \textit{(A semester is a time for students to complete certain courses of the curriculum. A regular semester consists of 15 weeks of instruction and 2 to 3 weeks for assessment activities (final exams, midterms, tests, etc.). A summer semester has a minimum of 5 weeks of instruction and 1 week for exams. Depending on the circumstances each year, the teaching plan for a semester may be adjusted by the decision of the Rector. 
        A school year consists of two regular semesters. Depending on the conditions, the University may organize additional summer semesters. The registration for summer semester courses is regulated in Article 14 of this regulation.)} \\ 
        \textbf{\textit{Question:}}\\
        Trường có những loại học kỳ gì? \textit{(What types of semesters does the University have?)} \\ 
        \textbf{\textit{Extractive answer wrong:}} \\
        Trường có\#học kỳ chính\#và\#học kỳ hè \textit{(the University have\#regular semester\#and\#summer semester)} \\ 
        \textbf{\textit{Extractive answer correct:}} \\
        học kỳ chính\#học kỳ hè \textit{(regular semester\#summer semester)} \\ 
        \textbf{\textit{Abstractive answer wrong:}}\\
        Trường có hai loại học kỳ: học kỳ chính và học kỳ hè. \textit{(The University has two types of semesters: regular semesters and summer semesters.)} \\
        \textbf{\textit{Abstractive answer correct:}}\\
        Trường có các loại học kỳ: học kỳ chính và học kỳ hè. \textit{(The university has semester types: regular semesters and summer semesters.)} \\ 
        \hline
    \end{tabular}
    \begin{tablenotes}
      \item Notes: The wrong extractive answer is highlighted with a red background, and the correct extractive answer is highlighted with blue text.
    \end{tablenotes}
    \label{tab:example_extractive_answer}
\end{table}

For reason types, our guidelines provide definitions and examples for reason types such as "Word-matching", "Paraphrasing", "Math", "Coreference", "Causal relation", and "Logic" based on the paper by Sugawara et al. (2018) \cite{sugawara-etal-2018-makes}. These types of reasoning are defined as follows:
\begin{itemize}
    \item \textbf{Word matching}: This involves exact word matching between words in the question and words in the context, or the answer connected to the question matches a sentence in the context.
    \item \textbf{Paraphrasing}: Questions rephrase the meaning of a context by altering vocabulary and grammar or using different knowledge to formulate the question.
    \item \textbf{Math}: Questions involving mathematics, where the answer requires applying mathematical operations or comparisons to solve the question.
    \item \textbf{Coreference}: This reasoning type involves answers that are entities. To identify these entities, one must refer to words or phrases in one or more different sentences that represent the entity being sought.
    \item \textbf{Causal relation}: The answer may explain the cause leading to the result mentioned in the question, or the question might inquire about the cause leading to the result mentioned in the answer.
    \item \textbf{Logic}: Utilizing knowledge from the context and the question to infer the answer, commonly seen in Yes/No questions.
\end{itemize}

We encourage creators to focus on creating questions that involve paraphrasing, math, coreference, causal relations, and logic. We do not encourage creators to create word-matching questions. Because word-matching is the easy reasoning type, a sufficient amount of training data can still achieve good results.


\subsubsection{Creator Agreement} \label{subsec:creator_agreement}

We have 7 creators, all university students from the same institution. These creators underwent training on the guidelines and performed multiple rounds of checks. The question-answer pairs must adhere to the guidelines, spelling, and structure of the sentence, ensuring diverse usage of reason types and question types. During each round of evaluation with 100 context-question pairs, creators must independently formulate answers. After that, creators will cross-check each other, provide feedback, and agree on answer writing in the regular meetings. We evaluated the similarity between the creators based on F1-score and BERTScore \cite{BERTScore} metrics. After three rounds of evaluations with 300 questions, the average results of the 7 creators will be presented as shown in Table \ref{tab:creator_agreement}.

\begin{table}[!ht]
\centering
\caption{The average similarity score of the 7 creators after 3 phases.}
\begin{tabular}{lcc}
\hline
\textbf{}   & \multicolumn{1}{l}{\textbf{F1 (\%)}} & \multicolumn{1}{l}{\textbf{BERTScore (\%)}} \\ \hline
Phase 1 & 56.00                       & 94.21                              \\
Phase 2 & 66.10                       & 94.49                              \\
Phase 3 & 68.36                       & 95.33                              \\ \hline
\end{tabular}

\label{tab:creator_agreement}
\end{table}

\subsubsection{Question-answer Creation} \label{subsec:question-answer_creation}

The dataset consists of 294 articles divided into two parts: Part 1 includes the first 146 articles labeled by 4 creators, while Part 2 comprises the remaining 148 articles labeled by the remaining 3 creators. This division of annotations and articles ensures diversity throughout the question-answer creation process. This approach allows us to maximize information extraction from the articles across different aspects while preventing duplication in question-answer pairs as in the case of all 7 people labeling all 294 contexts.

Each creator is required to generate at least 300 question-answer pairs in one week. The guidelines are strictly to ensure consistency across the dataset. We encourage creators to pose questions that involve challenging forms of inference, such as paraphrasing, inference from multiple sentences, and inference from a single sentence.

\subsubsection{Data Validation} \label{subsec:data_validation}

After each week of data creation, the creators will perform self-checks and cross-checks similar to the training phases in Section \ref{subsec:creator_agreement}. During the self-check process, each creator will review the question-answer pairs from the previous week and make corrections if any errors are found. In the cross-check process, each creator will examine the work of others to ensure adherence to the guidelines and identify errors in the data created by others. Throughout the cross-check process, we will review data from all creators to ensure no errors remain.

Upon completion of the cross-check process, we will hold discussions to address any issues encountered by the creators, propose solutions, and reach a consensus among all creators regarding these errors. In addition, we will update the guidelines weekly to address errors or exceptions.

In addition to the weekly evaluation and error correction processes, we will conduct a final review and error correction after completing the dataset in the last week to ensure consistency once again. Following this process, the dataset can be used for training and testing models.

\subsubsection{Data Splitting} \label{subsec:data_splitting}

After validating the data, we partitioned the data set into three subsets: training, development (validation), and testing, with an 8:1:1 ratio. The balanced allocation between the development and testing subsets is intended to ensure a fair and precise evaluation of the model.

\subsection{Dataset Analysis}

\subsubsection{Overall Statistics}

In this section, we conducted an overview analysis of the dataset regarding aspects such as the number of articles and the length of texts within the dataset. The ViRHE4QA dataset comprises 9,758 question-answer pairs from 294 articles within the domain of university training regulations. We conducted statistical analysis on the dataset regarding aspects such as the number of documents, number of articles, number of question-answer pairs, average word count{\footnote{We count words based on whitespace segmentation.}} in documents, articles, questions, extractive length, and abstractive length of the ViRHE4QA dataset, comparing these with the UIT-ViQuAD 1.0 dataset as shown in Table {\ref{tab:overview_corpus}}.

\begin{table}[htbp]
\centering
\caption{Overview statistics of the ViRHE4QA dataset.}
\label{tab:overview_corpus}
\resizebox{1\linewidth}{!}{
\begin{tabular}{lcccccccc}
\hline
 & \multicolumn{4}{c}{\textbf{ViRHE4QA}} & \multicolumn{4}{c}{\textbf{UIT-ViQuAD}} \\ \hline
 & \textbf{Entire} & \textbf{Train} & \textbf{Dev} & \textbf{Test} & \textbf{Entire} & \textbf{Train} & \textbf{Dev} & \textbf{Test} \\ \hline
Number of documents & 21 & 21 & 21 & 20 & - & - & - & - \\
Number of articles & 294 & 294 & 258 & 256 & 5,109 & 4,101 & 515 & 493 \\
Number of question-answer pairs & 9,758 & 7,806 & 976 & 976 & 23,074 & 18,579 & 2,285 & 2,210 \\
Average article length & 251.10 & 251.10 & 268.71 & 272.48 & 177.97 & 178.98 & 170.31 & 177.56 \\
Average name of document length & 16.84 & 16.86 & 16.78 & 16.71 & - & - & - & - \\
Average question length & 17.09 & 17.08 & 17.05 & 17.25 & 14.49 & 14.56 & 13.98 & 14.45 \\
Average extractive answer length & 24.18 & 23.89 & 26.49 & 24.19 & 10.14 & 10.02 & 10.49 & 10.82 \\
Average abstractive answer length & 35.16 & 34.85 & 37.72 & 35.05 & - & - & - & - \\ \hline
\end{tabular}}
\end{table}

Due to the close-domain dataset, the number of articles and question-answer pairs in ViRHE4QA is lower compared to the UIT-ViQuAD dataset. However, the average length of the articles in ViRHE4QA is longer than that of the UIT-ViQuAD dataset. Furthermore, the average length of the questions and the extractive answers in ViRHE4QA is higher than in UIT-ViQuAD. This poses a challenge for language models to locate and extract information accurately within longer contexts.

\subsubsection{Length-based Analysis}

To understand more about our dataset and domain, we performed statistics on the number of question-answer pairs grouped by ranges of article length (Table \ref{tab:article_length}), question length (Table \ref{tab:question_length}), and answer length (Table \ref{tab:answer_length}). Articles with lengths ranging from 101 to 256 words accounted for the largest proportion, with 3,422 question-answer pairs. However, it should be noted that articles with lengths less than 100 words had the smallest number of pairs, and articles longer than 512 words ranked second highest with 2,306 question-answer pairs. This poses a challenge in our dataset as most current language models accept a maximum input of 512 tokens.

\begin{table}[htbp]
\centering
\caption{Statistics on the number of question-answer pairs with the length of the article.}
\label{tab:article_length}
\begin{tabular}{ccccc}
\hline
\multicolumn{1}{l}{\textbf{Article length}} & \textbf{Entire} & \textbf{Train} & \textbf{Dev} & \textbf{Test} \\ \hline
\textless{}101 & 659 & 515 & 79 & 65 \\
\textbf{101-256} & \textbf{3,422} & \textbf{2,724} & \textbf{349} & \textbf{349} \\
257-400 & 2,019 & 1,639 & 189 & 191 \\
401-512 & 1,352 & 1,086 & 130 & 136 \\
\textgreater{}512 & 2,306 & 1,842 & 229 & 235 \\ \hline
\end{tabular}
\end{table}

Regarding question length, most are between 8 and 14 words, with a significant number also ranging from 15-21 words, showing relatively little difference compared to the 8-14 word range. Regarding the length of the answer, the highest proportion of extractive answers was less than 21 words, considerably more than other lengths. Meanwhile, abstractive answers predominantly fell within the 21-40 word range. This can be understood because of our guidelines, where extractive answers are expected to be the shortest answers, and abstractive answers represent a combination of the question and an extractive answer. This analysis shows that our dataset presents significant challenges regarding text length for current language models.

\begin{table}[htbp]
\centering
\caption{Statistics on the number of question-answer pairs with the length of the question.}
\label{tab:question_length}
\begin{tabular}{ccccc}
\hline
\multicolumn{1}{l}{\textbf{Question length}} & \textbf{Entire} & \textbf{Train} & \textbf{Dev} & \textbf{Test} \\ \hline
\textless{}8 & 414 & 338 & 43 & 33 \\
\textbf{8-14} & \textbf{3,699} & \textbf{2,951} & \textbf{382} & \textbf{366} \\
15-21 & 3,411 & 2,727 & 337 & 347 \\
22-28 & 1,515 & 1,224 & 134 & 157 \\
\textgreater{}28 & 719 & 556 & 80 & 73 \\ \hline
\end{tabular}
\end{table}

\begin{table}[!ht]
\centering
\caption{Statistics on the number of question-answer pairs with the length of the answers.}
\label{tab:answer_length}
\begin{tabular}{ccccccccc}
\hline
\multirow{2}{*}{\textbf{Length}} & \multicolumn{4}{c}{\textbf{Extractive answer}} & \multicolumn{4}{c}{\textbf{Abstractive answer}} \\ \cline{2-9} 
 & \textbf{Entire} & \textbf{Train} & \textbf{Dev} & \textbf{Test} & \textbf{Entire} & \textbf{Train} & \textbf{Dev} & \textbf{Test} \\ \hline
\textless{}21 & \textbf{6,622} & \textbf{5,314} & \textbf{645} & \textbf{663} & 3,401 & 2,740 & 325 & 336 \\
21-40 & 1,744 & 1,400 & 172 & 172 & \textbf{4,305} & \textbf{3,435} & \textbf{427} & \textbf{443} \\
41-60 & 581 & 451 & 63 & 67 & 1,044 & 841 & 103 & 100 \\
61-80 & 288 & 229 & 32 & 27 & 428 & 332 & 50 & 46 \\
\textgreater{}80 & 523 & 412 & 64 & 47 & 580 & 458 & 71 & 51 \\ \hline
\end{tabular}
\end{table}

\subsubsection{Type-based Analysis}

In this section, we conducted an analysis of the question types and the answer types in the test set (976 samples). To ensure accuracy, we manually classified the questions following the guidelines in section \ref{subsec:guidelines_creation}, which include 9 question types: What, Who, When, Where, Which, Why, How, Yes/No, and Others; and 6 reason types: Word-matching, Paraphrasing, Math, Coreference, Causal relation, and Logic.

\begin{figure}[!ht]
    \centering
    \includegraphics[width=0.7\linewidth]{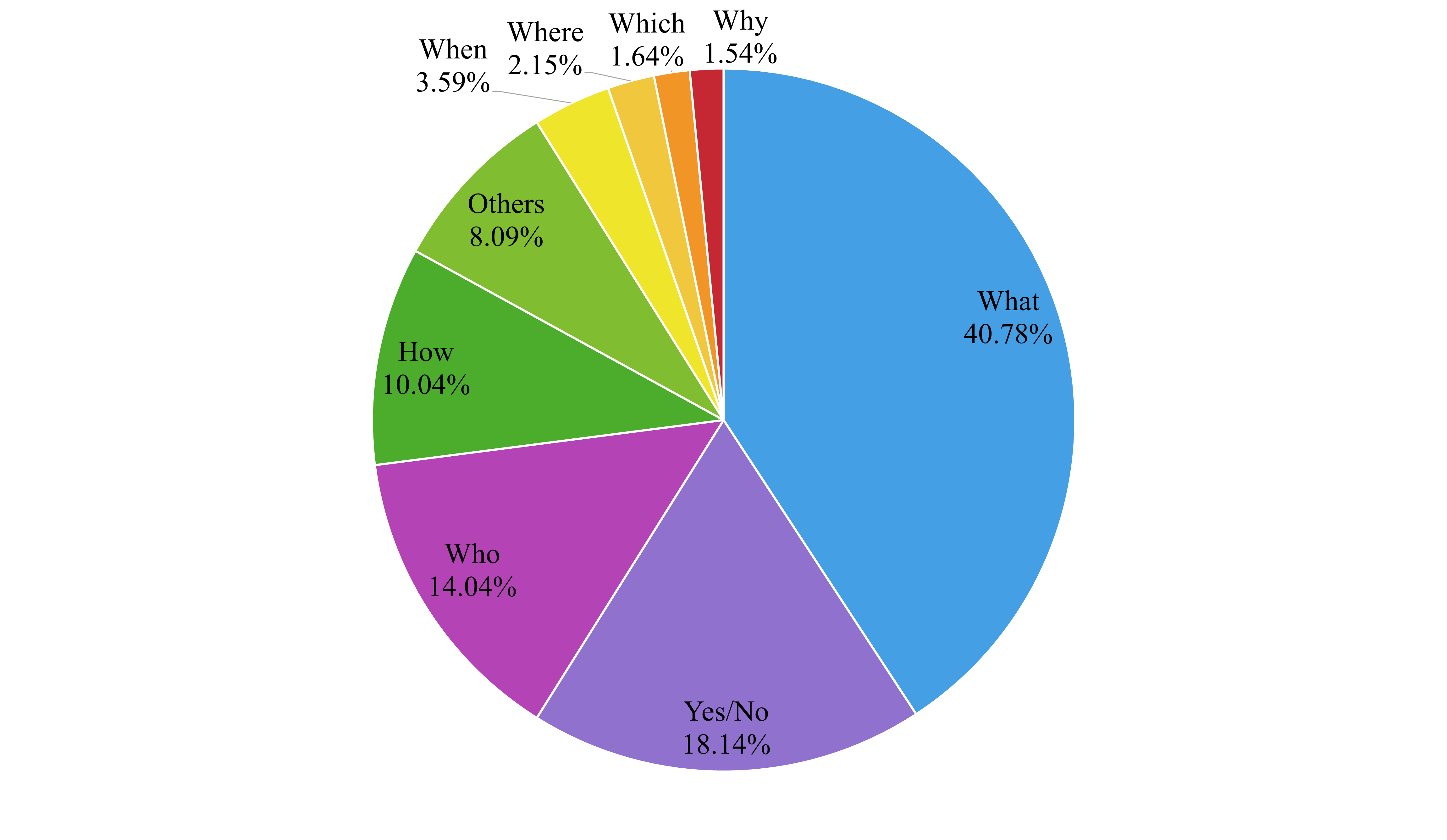}
    \caption{Distribution of question types in the ViRHE4QA dataset. We categorized questions into 9 types: What, Who, When, Where, Which, Why, How, Yes/No, and Others.}
    \label{fig:question_type_distribution}
\end{figure}

Figure \ref{fig:question_type_distribution} shows that the "What" type of question had the highest proportion at 40.78\%, followed by the "Yes/No" type at 18.14\%. Questions categorized as "When", "Where", "Which", and "Why" accounted for a very small proportion (together less than 10\%). Compared with the UIT-ViQuAD and UIT-ViNewsQA datasets, our dataset exhibits similar characteristics, with the "What" type of question being predominant (40.78\% compared to 49.97\% in UIT-ViQuAD and 54.35\% in UIT-ViNewsQA).

\begin{figure}[!ht]
    \centering
    \includegraphics[width=0.7\linewidth]{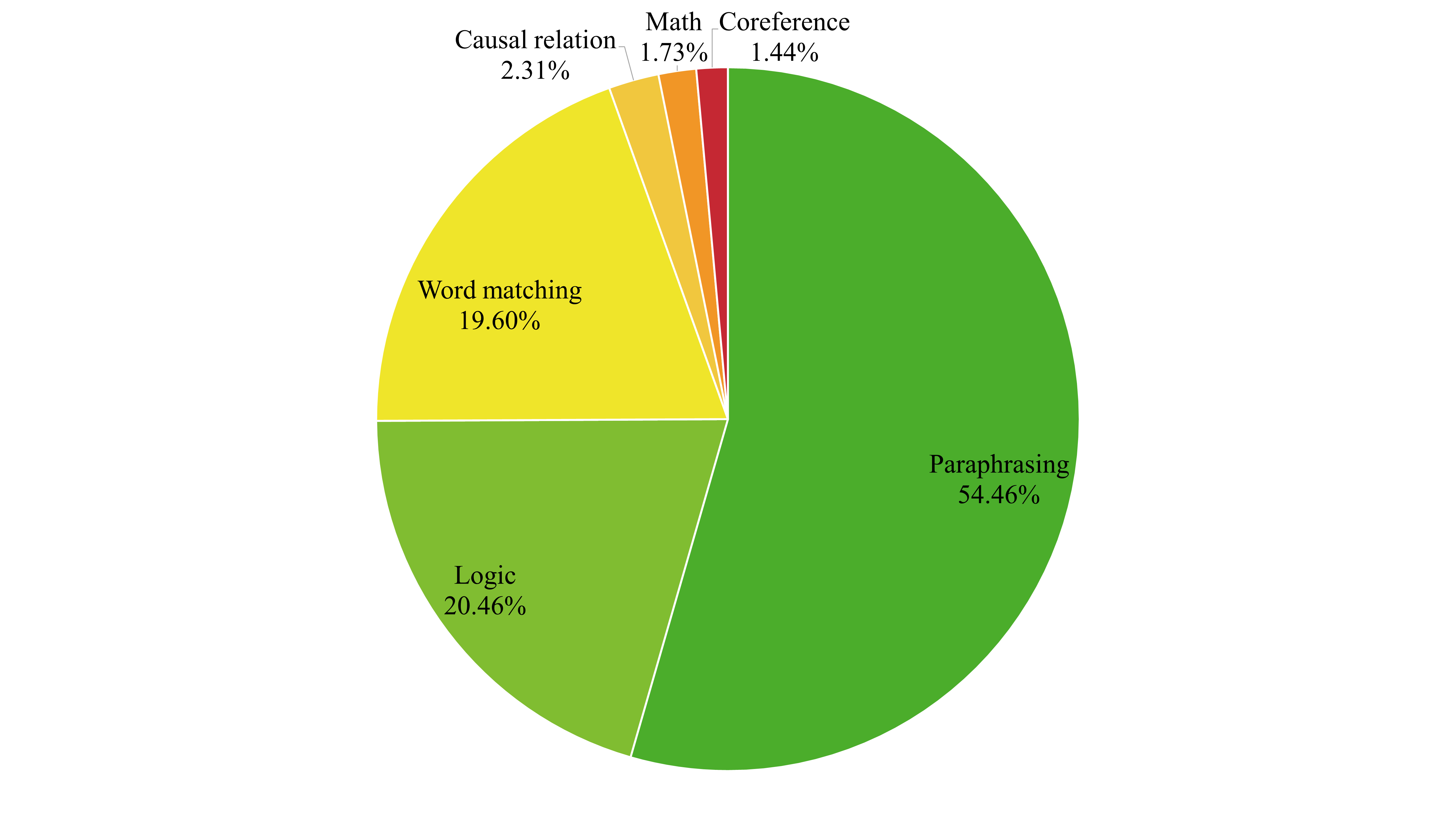}
    \caption{Distribution of reasoning types in the ViRHE4QA dataset. We categorized reasoning into 6 types: Word matching, Paraphrasing, Math, Coreference, Causal relation, and Logic.}
    \label{fig:reasoning_type_distribution}
\end{figure}

For reasoning types, according to Figure \ref{fig:reasoning_type_distribution}, Paraphrasing had the highest proportion at 54.46\%, followed by Logic at 20.46\%, and then Word-matching at 19.60\%. Math, Causal relation, and Coreference types had relatively low proportions at 5.48\% total. We request creators limit the use of word-matching question-answer formats to enhance diversity and challenge the dataset. Logical reasoning types are more prevalent because this type of reasoning is closely related to yes/no questions.

\section{Our Proposed Method}\label{QAS}
\begin{figure}[!ht]
    \centering
    \includegraphics[width=0.75\linewidth]{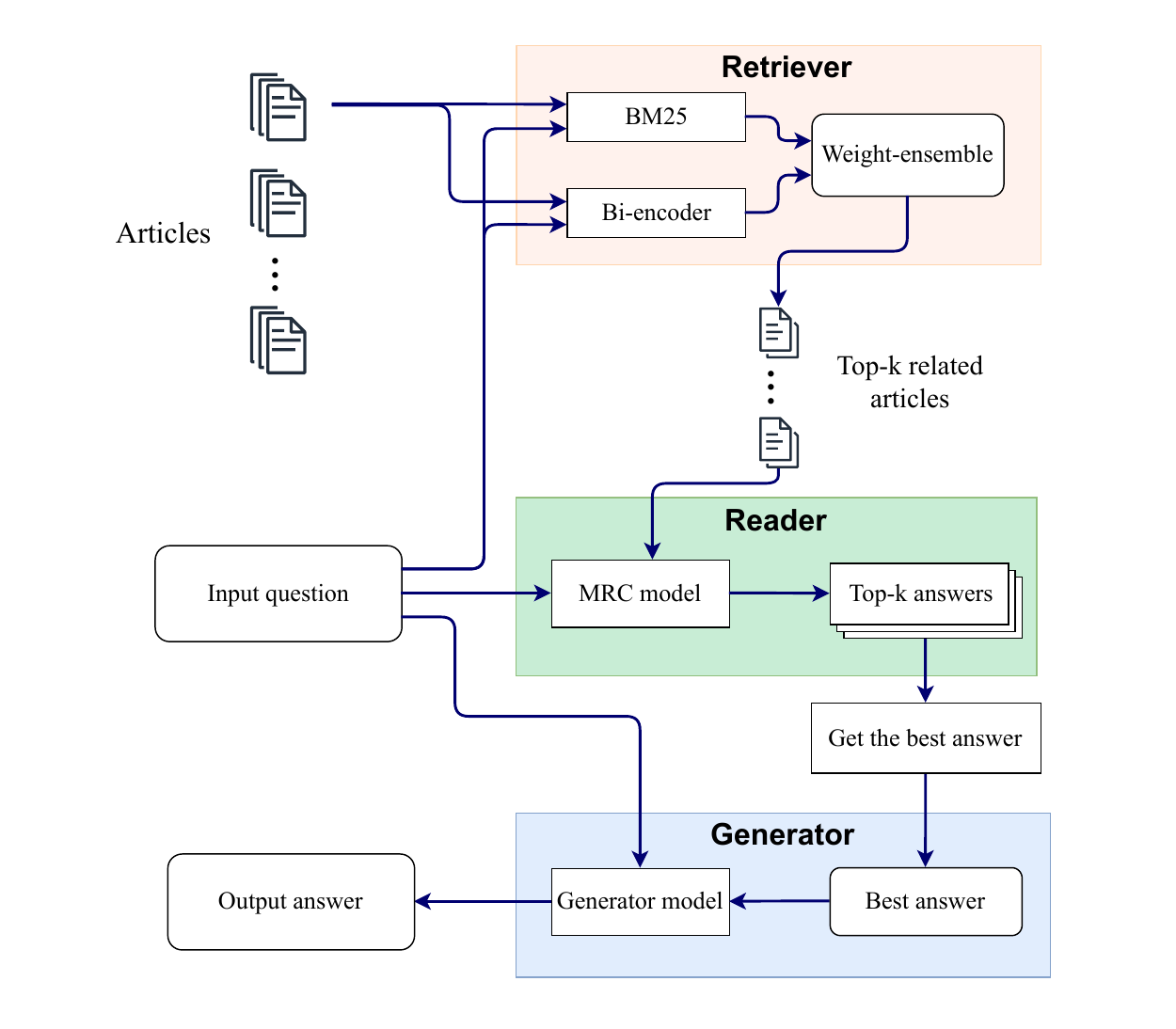}
    \caption{Diagram illustrates the R2GQA system consisting of three modules Retriever-Reader-Generator.}
    \label{fig:R2G-QA_system}
\end{figure}

In this section, we will present the question-answering system for abstract answers that we propose. This system consists of three modules: Document Retriever, Machine Reader, and Answer Generator. We named this system R2GQA, with an overall structure depicted in Figure \ref{fig:R2G-QA_system}.

\subsection{Document Retriever}

The Retriever module uses questions to retrieve contexts that contain answers or relevant information. These contexts are then fed into the machine reader module to extract answers. Additionally, the question scores corresponding to each context will be used to combine with the scores of the answers after the Reader module is executed to select the most accurate answer for the input question.
\subsubsection{Lexical Retrieval}
Retrieval methods based on lexical similarity employ the degree of overlap between a question and a document to determine relevance. BM25 and TF-IDF are two popular examples of this approach. However, these methods often fail when dealing with queries and documents that exhibit intricate semantic structures due to the limited extent of lexical overlap. This limitation arises from the inability of vectors to capture the true meaning of words.

\textbf{TF-IDF}: TF-IDF, which stands for "Term Frequency-Inverse Document Frequency" is a widely used technique in natural language processing (NLP) for preprocessing text data. This statistical method assesses the significance of a term within a document or dataset. TF-IDF is calculated by two factors: $tf(w, c)$ and $df(w, C)$.
\begin{equation}
    TF\text{-}IDF(w,c,C) = tf(w, c) \cdot idf(w, C)
\end{equation}
\begin{itemize}
\item TF (term frequency) is the frequency of occurrence of a word in a document. The TF value of a word $w$ in context $c$ is calculated according to the following formula:

\begin{equation} 
tf(w,c) = \frac{n(w,c)}{n(c)}
\end{equation}

$n(w, c) $ denotes the number of occurrences of the term $w$ in the context $c$.
$n(c)$ denotes the total number of occurrences of all terms in context $c$.
\item Inverse Document Frequency (IDF) is the inverse frequency of a term within a dataset. In the document collection, each term has a unique IDF value computed by the formula.
\begin{equation}
    idf(w, C) =  \log\frac{\mid C \mid}{\mid c_{w \in C}\in C \mid}
\end{equation}

$ \mid C \mid $ denotes the total number of documents in dataset $C$. $\mid c_{w \in C} \in C \mid$  is the number of documents $c$ that contain the term $w$ in the dataset. If the term does not appear in any context within the dataset $C$, the denominator would be 0, leading to an invalid division. Therefore, it is commonly replaced with the formula 1+ $\mid c_{w \in C} \in C \mid$.

\end{itemize}

TF-IDF transforms each document in a dataset into a vector representation, often referred to as document embedding. By combining the TF-IDF scores of each term in a document, a vector is formed that places the document within a high-dimensional space. This vector can be used as input for various machine learning models or for computing similarities between documents.

\textbf{BM25}: BM25 is a widely used ranking function in information retrieval to compute and rank the similarity between two texts. BM25 is a simple method commonly employed in question-answer tasks to search the context relevant to the input question. Similarly to TF-IDF, BM25 computes a score for each context in a dataset based on the frequency of the question terms in the context. BM25 also considers document length and term frequency saturation.
\begin{equation}
BM25(q, c) =  \sum_{i=1}^{ \mid q \mid } IDF(q_i) \cdot \frac{f(q_i, c) (k + 1)} {f(q_i, c) + k - (1 - b + b \cdot \frac{ \mid c \mid }{C_{avgl}})}
\end{equation}
Where:
\begin{itemize}
    \item $q$ represents the question.
    \item $c$ signifies a context within the dataset.
    \item $|q|$ indicates the question length.
    \item $IDF(q_i)$ stands for the Inverse Document Frequency of the $i$-th question ($q_i$).
    \item $f(q_i, d)$ is the frequency of the $i$-th question ($q_i$) within the context $c$.
    \item $k$ and $b$ are adjustable parameters used in certain weighting frameworks.
    \item $|c|$ denotes the length of context $c$.
    \item $C_{avg}$ refers to the average context length within the dataset.
\end{itemize}

The BM25 formula diverges from TF-IDF in several significant aspects. Firstly, it employs a non-linear approach to calculate term frequency weights, which leads to an exponential increase in weighting with higher term frequencies. Secondly, it normalizes context lengths based on specific terms, decreasing the weighting of frequently occurring terms in extensive contexts. Additionally, the parameters $k$ and $b$ are tunable to adjust the emphasis on term frequency and context length normalization in the calculation.

\subsubsection{Contextualized-based Retrieval} \label{thesis_be}
\textbf{Bi-Encoder:} Reimers and Gurevych \cite{sbert} proposed the Bi-Encoder in 2019. Bi-Encoder are utilized across various tasks such as NLI, STS, Information Retrieval, and Question Answering systems. For the question-answer system task, the contexts in the dataset are encoded independently into vectors. The input question is then encoded and embedded in the vector space of the contexts to compute similarity scores with each context. Based on these scores, relevant contexts related to the question can be determined. 

One of the training methods for bi-encoders involves using the MarginMSE loss function. MarginMSE is based on the paper of Sebastian et al. (2020) \cite{marrgin}. Similarly to MultipleNegativesRankingLoss, to train with MarginMSE, triplets (question, context 1, context 2) are required. However, unlike MultipleNegativesRankingLoss, context 1 and context 2 do not need to strictly be positive/negative; both can be relevant or irrelevant to a given question. 

For training the bi-encoder with MarginMSE, the following procedure is undertaken: First, scores are computed for each pair (question, context 1) and (question, context 2). The distance of score ($ScoreDistance$) between the two pairs serves as the label for the triplet (question, context 1, context 2).  The $ScoreDistance$ is calculated using the formula: 
\begin{equation}
    ScoreDistance = Score_{(question, context 1)} -  Score_{(question, context 2)}
\end{equation}

In training the bi-encoder, question, context 1, and context 2 are encoded into vector spaces, and then the score of (question, context 1) and (question, context 2) is computed. Subsequently, the BDistance is computed by subtracting the score of (question, context 2) from the score of (question, context 1). The purpose of training is to optimize the error between ScoreDistance and BDistance.

\subsubsection{Lexical-Contextual Retrieval} 
\textbf{Weight Ensemble}: We combined the scores of each context when querying by lexical and Bi-Encoder using a weight $\alpha$ for each top\_k. The scores calculated from the bi-encoder model were normalized to values in the range [0; 1]. After combining, we extracted the top\_k contexts with the highest scores. The combination formula is as follows:
\begin{equation} \label{fomula_weight}
    Score=
    ScoreBM25_{(\textit{question, context})} \cdot \alpha + (1-\alpha) \cdot ScoreBE_{(\textit{question, context})}
\end{equation}

\textbf{Multiplication Ensemble}: We calculated the product of the scores from each context calculated by TF-IDF and Bi-Encoder. Similarly to the weight ensemble, the scores calculated from the Bi-Encoder model were normalized to values in the range [0, 1]. Finally, we extracted the top\_k contexts with the highest scores.
\begin{equation} \label{fomula_multi}
    Score_{(\textit{question, context})} = ScoreBM25_{(\textit{question, context})} \cdot ScoreBE_{(\textit{question, context})}
\end{equation}
We retrieve contexts using a combined method through the following main steps: First, we score the contexts in the database using the BM25 method. Second, we retrieve the scores of the contexts using the Bi-Encoder method and normalize them to the range [0,1]. Third, we obtain the combined scores of the contexts using the combined method. Finally, we extract the top\_k highest scores and their corresponding contexts. Algorithm \ref{ensemble_algorithm} details our combined querying process.

\begin{algorithm}
\caption{Query the top\_k most relevant contexts from the input question.}\label{ensemble_algorithm}
\begin{algorithmic}[1]
    \State \textbf{Input:} Question $Q$.
    \State \textbf{Output:} List of top\_k contexts relevant to question $Q$ with the highest scores and corresponding scores.
\Function{QueryContext}{Question $Q$}
    \State $TC \gets$ List of contexts tokenized.
    \State $bm25 \gets$ BM25 function.
    \State $tokenized_q \gets$ Question tokenized by pyvi. 
    \State $ScoresBM25 \gets$ List of scores of contexts from bm25($Q$, $TC$).
    
    \State $q \gets$ Extracted contextual vector encoded by bi-encoder ($Q$).
    \State $TCE \gets$ List of contextual vectors encoded by bi-encoder($TC$).
    \State $ScoresBE \gets$ List of scores of contexts from similarity($Q$, $TC$).
    \State $SScoresBE \gets$ List of scores normalized to the range [0; 1] of contexts with the question.
    \State $ScoresEnsemble \gets$ List of combined scores of contexts from $ScoresBM25$ and $SScoresBE$ (Combination formula is either formula \ref{fomula_weight} or formula \ref{fomula_multi}).
    \State $ScoresEnsemble \gets$ Array of scores ranked in descending order.
    \State $KContexts \gets$ List of top\_k contexts with the highest scores for the input question based on the scores of the $ScoresEnsemble$ array and corresponding scores.
    \State \Return $KContexts$
\EndFunction
\end{algorithmic}
\end{algorithm}

\subsection{Machine Reader} \label{reader}
\begin{figure}[!ht]
    \centering
    \includegraphics[width=0.75\textwidth]{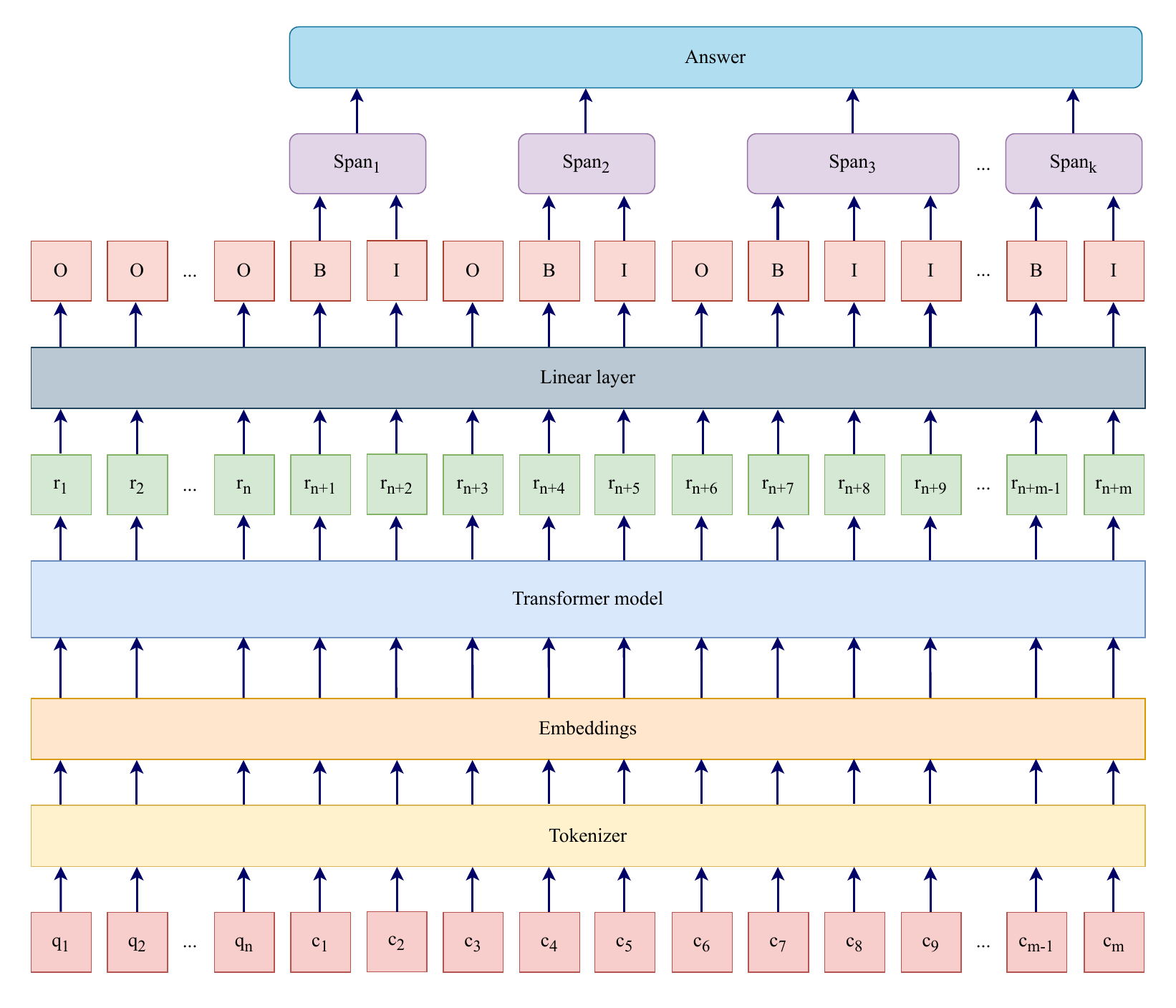}
    \caption{The sequence tagging approach for the Reader module. $ q_{\text{1}} $, $ q_{\text{2}} $,... $ q_{\text{n}} $ are the words of the question, $ c_{\text{1}} $, $ c_{\text{2}} $, $ c_{\text{3}} $,..., $ c_{\text{n}} $ are the words of the context. $ r_{\text{1}} $, $ r_{\text{2}} $, $ r_{\text{3}} $,..., $ r_{\text{m+n}} $ are the contextualized representations of the input words, and $ span_{\text{1}} $, $ span_{\text{2}} $,..., $ span_{\text{k}} $ are the spans in extractive answer.}
    \label{fig:BIO_approach}
\end{figure}

In this study, we implement a Reader module based on the sequence tagging approach - BIO format (B - beginning, I - inside, O - outside). The BIO approach means that tokens in the input will be classified into B, I or O labels. If a token is labeled B or I, it means that the token is part of the answer; otherwise, it does not appear in the answer. This approach is commonly used as a method for extracting answers for extractive reading comprehension tasks \cite{multispanqa,simple}. Figure \ref{fig:BIO_approach} illustrates this approach in detail.

There are various methods for training models using the BIO approach. In recent years, transfer learning methods have proven to be effective for MLP tasks due to their pre-training on large datasets. For machine reading comprehension tasks, several state-of-the-art (SOTA) high-performance models have been trained on multilingual datasets, such as multilingual BERT (mBERT) \cite{devlin-etal-2019-bert} and XLM-RoBERTa \cite{xlmr}. Currently, there are also models specifically designed for Vietnamese, such as CafeBERT \cite{cafebert}, ViBERT \cite{ViBERT}, and vELECTRA \cite{ViBERT}. Therefore, we utilize these models to implement the Reader module.

\begin{enumerate}[1.]
   \item \textbf{XLM-RoBERTa} \cite{xlmr} is a pre-trained multilingual language model. It was trained on the CommonCrawl dataset, which includes text data from over 100 languages (including more than 137GB of Vietnamese text data). XLM-RoBERTa comes in two versions: large (with 24 layers) and base (with 12 layers).

    \item \textbf{CafeBERT} \cite{cafebert} is a language model built on top of XLM-RoBERTa. This model was trained on approximately 18GB of Vietnamese text data. CafeBERT outperforms XLM-RoBERTa on the VLUE benchmark \cite{cafebert} (including the reading comprehension task).

    \item \textbf{vELECTRA} \cite{ViBERT} is a model trained on a massive dataset of Vietnamese text data, reaching 58.4GB in size. The authors used BERT as the foundation for the generator module and ELECTRA for the discriminatory module.

     \item \textbf{ViBERT} \cite{ViBERT} is a model trained on 10GB of Vietnamese text data. Its architecture is based on the BERT model. In addition to using similar layers to BERT, the authors added two layers before the final linear layer: a bidirectional RNN layer and an Attention layer. This results in ViBERT having a total of 5 layers.
\end{enumerate}

\subsection{Answer Generator}

\begin{figure}
    \centering
    \includegraphics[width=0.75\linewidth]{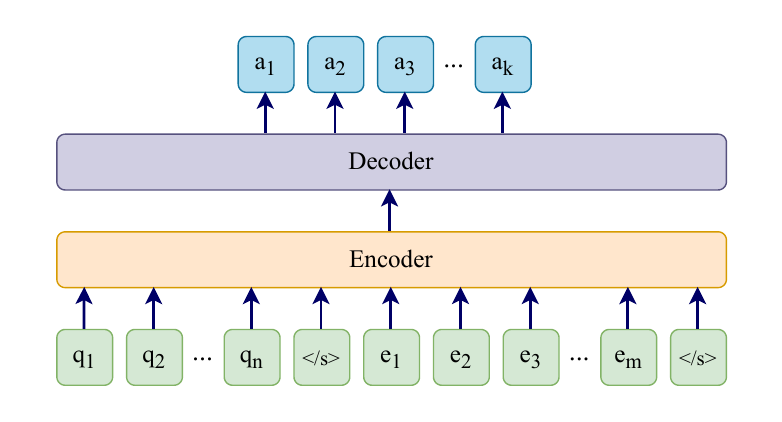}
    \caption{The structure of the module Generator. $ q_1, q_2, \ldots, q_n $ are the words in the question, $ e_1, e_2, e_3, \ldots , e_m$ are the words in the extractive answer. $ a_1, a_2, a_3, \ldots, a_k $ are the outputs of the module Generator.}
    \label{fig:generator module}
\end{figure}

In the Retrieval-Reader-Generator system, the Generator module operates at the final stage of the answer generation process. The main function of this module is to merge the information from the question and the extractive answer. This module uses a sequence-to-sequence structure, often seen in tasks such as machine translation or summarization, shown in Figure \ref{fig:generator module}. This helps generate a complete, human-like answer.

Mathematically, the function $ f $ representing the Generator module is expressed as follows:
\begin{equation}
    A_{abstractive} = f ( Q, A_{extractive})
\end{equation}

Here, the inputs (Q, $ A_{extractive} $) represent:
\begin{itemize}
    \item $ Q $: the question.
    \item $ A_{extractive} $: the extractive answer taken from the Reader module.
\end{itemize}

The output $ A_{abstractive} $ is the generated answer, refined, coherent, and synthesizes information from the question and the extracted context in a more understandable form.

Vietnamese language generator models are evolving, employing transfer learning methods to enhance performance. In this paper, we use state-of-the-art (SOTA) generator models for Vietnamese, including multilingual models such as mBART-50 \cite{mbart50}, mT5 \cite{mt5}; and monolingual models such as BARTpho \cite{bartpho} and ViT5 \cite{vit5}.

\begin{enumerate}[1.]
    \item \textbf{mBART-50} \cite{mbart50}: mBART-50 is an extension of BART (Bidirectional and Auto-Regressive Transformers), supporting multiple languages, including 137.3GB of Vietnamese data. mBART-50 is a denoising autoencoder trained with masked language modeling and permutation language modeling objectives. It has shown strong performance in various language generation tasks, such as translation and summarization.
    \item \textbf{mT5} \cite{mt5}: mT5 is an adaptation of T5 model for multilingual text generation tasks. The key to the innovation of mT5 lies in its ability to perform diverse Natural Language Processing (NLP) tasks within a unified framework, including text generation, translation, summarization, question answering, and more. This unified architecture simplifies NLP application deployment across languages, making it valuable for global communication and multilingual content generation. mT5 is trained on the mC4 dataset, including Vietnamese with 116B tokens, 79M pages, and constituting 1.86\% of the training data for the mT5 model.
    \item \textbf{BARTpho} \cite{bartpho}: BARTpho utilizes a "large" architecture and pre-training scheme similar to sequence-to-sequence denoising autoencoder of BART. It leverages a Vietnamese dataset of 20GB from PhoBERT, showing improved performance over mBART in Vietnamese text summarization tasks. BARTpho has two versions: $BARTpho_{word}$ and $BARTpho_{syllable}$, with the word version performing better.
    \item \textbf{ViT5} \cite{vit5}: ViT5 is a monolingual model developed for Vietnamese based on the T5 structure. It is built on 138GB of Vietnamese data from the CC100 dataset and is trained for Vietnamese abstractive summarization and Named Entity Recognition tasks. ViT5 significantly improves over current SOTA models in Vietnamese text summarization and competitive results in NER tasks.
\end{enumerate}

\section{Experiments and Results}\label{Ex}
\subsection{Metrics}
\subsubsection{P@k}
To evaluate the performance of the retrieval methods, we use the P@k measure. P@k measure is commonly used in information retrieval tasks, and some works such as XLMRQA \cite{XLMRQA}, SPBERTQA \cite{SPBERT}, LegalCQA \cite{legalcqa} have utilized it. In formula \ref{p@k}, P@k is the proportion of questions for which the relevant corresponding context appears in the contexts returned by the retrieval module. C$_{i_{pos}}$ is the relevant context corresponding to question q$_{i}$, and C$_{k}$(q$_{i}$) are the contexts returned by the retrieval module corresponding to question q$_{i}$. n is the number of questions.
\begin{equation} \label{p@k}
   P@k = \frac{1}{n}\sum_{i=1}^{n}
\begin{cases}
 1 & \text{ if } C_{i_{pos}} \in C_{k}(q_{i})  \\ 
 0 & \text{ if } C_{i_{pos}} \notin C_{k}(q_{i})  
\end{cases}
\end{equation}

\subsubsection{F1}
The F1-score is a widely used metric in natural language processing and machine reading comprehension. Evaluates the accuracy of the predicted answers by comparing individual words with those in the correct answers. The F1-score measures the overlap in words between the predicted answers and the ground-truth answers.
\begin{equation}
    \text{Precision} =  \frac{\text{the number of overlap words} }{\text{the total number of tokens in the predicted answer} }
\end{equation}
\begin{equation}
    \text{Recall} =  \frac{\text{the number of overlap words} }{\text{the total number of tokens in the gold answer} }
\end{equation}

\begin{equation}
    \text{F1-Score} = 2 \cdot \frac{\text{Precision} \cdot \text{Recall}}{\text{Precision} + \text{Recall}}
\end{equation}

\subsubsection{BLEU}
BLEU (Bilingual Evaluation Understudy) \cite{bleu} is a scoring method to measure the similarity between two texts in machine translation. BLEU compares contiguous word sequences in the machine-generated text with those in the reference text, counting matching n-grams with weighted precision. These matches are position independent. BLEU is described by the following formula:

\begin{equation}
BLEU_{\text{Score}} = \text{BP} \times \exp\left( \sum_{i=1}^{N} (w_i \cdot \log(p_i)) \right)
\end{equation}
Where:
\begin{itemize}
    \item BP (Brevity Penalty) is a brevity penalty factor to account for shorter translations compared to the reference translations.
    \item $exp$ denotes the exponential function.
    \item $\sum_{i=1}^{N} (w_i \cdot \log(p_i))$ represents the weighted sum of the logarithm of precisions $p_i$, where $w_i$ is the weight for the n-gram precision of order $i$, and $N$ is the maximum n-gram order considered in the calculation.

\end{itemize}

\subsubsection{ROUGE}
In addition to comparing model outputs directly, we assess their agreement by measuring the overlap in content. To do this, we leverage the ROUGE framework (Recall-Oriented Understudy for Gisting Evaluation) \cite{lin-2004-rouge}. ROUGE metrics are popular tools for automating text summarization and machine translation and analyze both the structure and vocabulary of the generated text compared to a reference answer. This study utilizes several ROUGE metrics, including:

\begin{enumerate}[1.]
    \item ROUGE-N: This metric focuses on counting matching sequences of words (n-grams) between the answer of the system and the ideal answer. Higher ROUGE-N scores indicate a greater degree of overlap in wording.
    \item ROUGE-L: This metric prioritizes finding the longest string of words that appears in the same order in both the predicted answer and the gold answer. It emphasizes the importance of word order compared to ROUGE-N.
\end{enumerate}

\subsubsection{BERTScore}
BERTScore \cite{BERTScore} is a metric used to evaluate the performance of text generation models, including machine translation and text summarization. This metric leverages the contextual understanding ability of language models to encode predicted answers and gold answers into embedding vectors and then computes the cosine similarity between these embeddings to provide a score for the quality of the generated text. The higher the score, the greater the similarity, indicating better performance of the answers of model.

BERTScore focuses on assessing semantic similarity rather than just lexical similarity like traditional metrics. This helps to evaluate the overall quality of text generation models more comprehensively. Additionally, BERTScore is available for multiple languages, allowing cross-lingual evaluation of text generation models.
\subsection{Experimental Design}
In this section, we provide detailed configurations of the three modules in R2GQA system: Document Retriever, Machine Reader, and Answer Generator. We conducted all experiments on the RTX 3090 GPU with 24GB VRAM from VastAI\footnote{\url{https://vast.ai/}}.
\subsubsection{Document Retriever}
The purpose of the Document Retriever module is to question for context that may contain answers to the questions.  We assign IDs to the contexts, which are used to map to the IDs of the contexts with the highest retrieval scores returned after performing the retrieval methods. For the Retriever module, we conduct experiments with 3 methods: lexical retrieval, contextual retrieval, and lexical-contextual retrieval with \textbf{top\_k} = [1, 5, 10, 15, 20, 25, 30].

 \textbf{Lexical retrieval}: We experiment with TF-IDF and BM25 methods. To enhance performance, we apply word segmentation using the Pyvi library \footnote{\url{https://pypi.org/project/pyvi/0.0.7.5/}} when conducting query experiments with TF-IDF and BM25.

 \textbf{Contextual retrieval}: We employ 2 approaches. In both approaches to contextual retrieval, we utilize word segmentation with Pyvi. \textbf{LME}: We only use pre-trained models (ViEmb\footnote{\url{https://huggingface.co/dangvantuan/vietnamese-embedding}}, ViSBERT\footnote{\url{https://huggingface.co/keepitreal/vietnamese-sbert}}, ViSimCSE\footnote{\url{https://huggingface.co/VoVanPhuc/sup-SimCSE-VietNamese-phobert-base}}, ViBiEncoder\footnote{\url{https://huggingface.co/bkai-foundation-models/vietnamese-bi-encoder}}) from Huggingface to encode the question and context, then employ cosine similarity to calculate similarity scores and re-rank based on these scores. \textbf{Bi-Encoder} \cite{sbert}: We continue to fine-tune pre-trained models from Huggingface with our data using the MarginMSE loss function. We train with epochs = 10, batch\_size = 24. The score for each pair, Score(question, context1), Score(question, context2) in Section \ref{thesis_be}, is computed by BM25 instead of Cross-Encoder.

 \textbf{Lexical-Contextual retrieval}: 
For both combination methods in Section \ref{thesis_be}, we experimented using an exhaustive search of $\alpha$ values in the range [0.1; 0.9] with a step size of 0.1. Through this process, we determined the $\alpha$ value with the highest performance. For the combination with TF-IDF, the highest performance was at $\alpha = 0.3$, and for BM25, it was $\alpha = 0.1$.

\subsubsection{Machine Reader}
For the Reader models, we implemented experiments with models and approaches as in Section \ref{reader}. The models were trained with epochs = 5, batch\_size = 8, learning\_rate = 5e-5, max\_seq\_length = 512. The optimizer used was AdamW. The evaluation metrics used were F1, BLEU1, and BERTScore.

Our data contains many contexts longer than 512 tokens, while the maximum length of the models we experimented with is 512 tokens. Therefore, we split each context longer than 512 tokens into multiple input features. To minimize information loss and preserve the semantics of the input features, we use the stride hyperparameter to create overlapping segments between two input features.

\subsubsection{Answer Generator} \label{experiment_generator}
We use the following generator models with specific configurations: mBART-large-50, mT5-base, ViT5-base and $BARTpho_{word}$. We added the token </s> to the model input to separate the question and extractive answer in the format \textit{Question </s> Extractive answer </s>}. For the BARTpho model, the input was formatted as \textit{<s> Question </s></s> Extractive answer </s>}. We perform word segmentation on $BARTpho_{word}$ using VncoreNLP before training the model. The model parameters were set as similarly as possible with epoch = 5, learning rate = 4e-05, max\_seq\_length = 1024 (512 for mT5 due to model limitations), batch\_size = 2, and using the AdamW optimizer. The metrics used in this section included: BLEU1, BLEU4, ROUGE-L, and BERTScore.

\subsection{Experiment Results}

This section initially assesses the performance of our Document Retriever and Machine Reader modules independently. Subsequently, it details experiments involving their integration into the R2GQA system, which is applied to close-domain question answering concerning legal regulations in higher education.

\subsubsection{Document Retriever} \label{retriever_result_section}

Based on Table \ref{retrieval_result}, it can be seen that the lexical query method combined with the contextual method produces the highest results for all values of the top\_k. The LME method consistently produces the lowest results as it has not been trained to understand context within our data domain. Comparing the two ensemble methods, we can observe that the weighted combination method between BM25 and Bi-Encoder provides the highest results for 5 out of the 7 top\_k values tested. Thus, it can be concluded that this method has the highest stability. Therefore, we will use this method for our end-to-end system (ViEmb\footnote{\url{https://huggingface.co/dangvantuan/vietnamese-embedding}}, ViSBERT\footnote{\url{https://huggingface.co/keepitreal/vietnamese-sbert}}, ViSimCSE\footnote{\url{https://huggingface.co/VoVanPhuc/sup-SimCSE-VietNamese-phobert-base}}, ViBiEncoder\footnote{\url{https://huggingface.co/bkai-foundation-models/vietnamese-bi-encoder}}).

\begin{table}[!ht] 
\caption{Result for each method in the Document Retriever module.}\label{retrieval_result}
\resizebox{1\linewidth}{!}{\begin{tabular}{cccccccccccc}
 \cline{1-12}
\multicolumn{1}{l}{\multirow{2}{*}{\textbf{Top\_k}}} & \multicolumn{1}{l}{\multirow{2}{*}{\textbf{TFIDF}}} & \multicolumn{1}{l}{\multirow{2}{*}{\textbf{BM25}}} & \multicolumn{4}{c}{\textbf{LME}} & \multicolumn{1}{l}{\multirow{2}{*}{\textbf{Bi-Encoder}}} & \multicolumn{2}{c}{\textbf{Weight-Ensemble}}          & \multicolumn{2}{c}{\textbf{Multiplication-Ensemble}}   \\ \cline{4-7} \cline{9-12} \multicolumn{1}{l}{} & \multicolumn{1}{l}{}                    & \multicolumn{1}{l}{}                               & \multicolumn{1}{l}{\textbf{(1)}} & \multicolumn{1}{l}{\textbf{(2)}} & \multicolumn{1}{l}{\textbf{(3)}} & \multicolumn{1}{l}{\textbf{(4)}} & \multicolumn{1}{l}{}                            & \multicolumn{1}{c}{TFIDF} & \multicolumn{1}{c}{BM25} & \multicolumn{1}{c}{TFIDF} & \multicolumn{1}{c}{BM25} \\ \hline
1                                           & 55,84                 & 68,75                 & 44,06 & 44,57 & 43,24 & 52,36 & 61,27                       & 64,75            & 72,13            & 63,93                & \textbf{72,74}                \\
5                                           & 86,66                  & 91,39                 & 74,49 & 73,98 & 72,34 & 76,85 & 89,86                       & 91,08            & \textbf{93,44}            & 90,06                & \textbf{93,44}                \\
10                                          & 93,55                  & 94,77                 & 82,17 & 84,53 & 82,58 & 84,43 & 94,67                       & 95,38            & \textbf{96,72}            & 95,08                & 96,52                \\
15                                          & 95,80                  & 96,93                 & 87,60 & 88,42 & 86,17 & 88,22 & 96,21                       & 96,82            & \textbf{97,44}            & 96,82                & 97,33                \\
20                                          & 97,34                  & 97,54                 & 91,09 & 90,68 & 89,45 & 90,16 & 97,03                       & 97,85            & \textbf{98,05}            & 97,95                & \textbf{98,05}                   \\
25                                          & 98,16                  & 98,05                 & 92,32 & 92,32 & 91,09 & 91,19 & 97,54                       & 98,16            & \textbf{98,36}            & 98,77                &\textbf{98,36}               \\
30                                          & 98,67                  & 98,16                 & 94,06 & 94,06 & 92,52 & 92,52 & 97,85                       & 98,57            & 98,46            & \textbf{98,87}                & 98,66                \\ \hline
\end{tabular}}
\end{table}

\subsubsection{Machine Reader} \label{reader_result_section}

Table \ref{tab:main_result_reader} shows that the XLM-RoBERTa-Large model achieves the best results on most metrics and answer types. The second-best-performing model is CafeBERT. Across the entire ViRHE4QA test dataset, XLM-RoBERTa-Large outperforms CafeBERT by 0.04\% on the F1 metric and 0.5\% on BLEU1, while CafeBERT surpasses XLM-R-Large by 0.69\% on BERTScore. However, overall, the XLM-R-Large model demonstrates more consistent results, outperforming CafeBERT on 2 out of 3 metrics. The ViBERT model performs the worst in all metrics and answer types, showing a significant gap compared to the other models.

\begin{table}[!ht]
\centering
\caption{Results of Reader models on our test set.} 
\label{tab:main_result_reader}
\resizebox{1\linewidth}{!}{
\begin{tabular}{lccccccccc} 
\hline
\multicolumn{1}{l}{\multirow{2}{*}{}} & \multicolumn{3}{c}{\textbf{All}} & \multicolumn{3}{c}{\textbf{1 span}} & \multicolumn{3}{c}{\textbf{> 1 span}} \\ 
\cline{2-10}
\multicolumn{1}{l}{} & \textbf{F1} & \textbf{BLEU1} & \textbf{BERTScore} & \textbf{F1} & \textbf{BLEU1} & \textbf{BERTScore} & \textbf{F1} & \textbf{BLEU1} & \textbf{BERTScore} \\ 
\hline
\textbf{XLM-R-Large} & \textbf{72.96} & \textbf{66.08} & 84.86 & 73.47 & \textbf{66.75} & 84.87 & \textbf{60.32} & \textbf{49.37} & \textbf{84.81} \\
CafeBERT & 72.92 & 65.58 & \textbf{85.55} & \textbf{73.49} & 66.25 & \textbf{85.68} & 58.87 & 49.10 & 82.44 \\
vELECTRA & 60.79 & 52.49 & 78.61 & 60.92 & 52.72 & 78.58 & 57.70 & 46.72 & 79.33 \\
XLM-R-Base & 60.68 & 53.07 & 76.40 & 61.05 & 53.53 & 76.35 & 51.55 & 41.65 & 77.44 \\
ViBERT & 55.26 & 47.56 & 73.71 & 55.60 & 48.00 & 73.83 & 46.86 & 36.71 & 70.46 \\ 
\hline
\end{tabular}}
\end{table}

Questions with single-span (1 span) answers achieve significantly better results than questions with multi-span (> 1 spans) answers across all models and metrics. For the XLM-R-Large model, performance on single-phrase answers exceeds that on multi-phrase answers by 13.15\%, 17.38\%, and 0.06\% on F1, BLEU, and BERTScore, respectively. This significant difference indicates that finding answers to questions where the answer appears in multiple locations in the context is much more challenging than when the answer is located in a single position.
\subsubsection{Answer Generator} \label{generator_result_section}

\begin{table}[!htbp]
\centering
\caption{Results of the Generator module with input are a Question and an Extractive answer.}
\label{tab:generator_module}
\begin{tabular}{lcccc}
\hline
\textbf{Model} & \textbf{BLEU1} & \textbf{BLEU4} & \textbf{BERTScore} & \textbf{ROUGE-L} \\ \hline
BARTpho & \textbf{81.83} & 70.49 & 94.35 & 85.79 \\
mBART & 81.82 & \textbf{73.21} & \textbf{95.20} & \textbf{86.35} \\
ViT5 & 80.15 & 72.04 & 94.52 & 85.70 \\
mT5 & 70.52 & 61.16 & 87.38 & 83.90 \\ \hline
\end{tabular}
\end{table}

According to the results in Table \ref{tab:generator_module}, the BARTpho model achieved the highest score on the BLEU1 metric, but mBART outperformed the remaining three metrics, including BLEU4, BERTScore, and ROUGE-L. Compared to mBART, the BARTpho and ViT5 models exhibited slightly lower performance, ranging from 1\% to 3\%. However, the mT5 model performed significantly worse than the other three models. This could be attributed to the mT5 model having an input length limit of only 512 tokens, while the other three models accept input lengths of up to 1024 tokens. This limitation notably affects the performance, especially with datasets containing contexts longer than 512 tokens, such as ViRHE4QA.

\subsubsection{End-to-End System}

\begin{table}[!ht]
\centering
\caption{Results of R2GQA system on the top\_k.} \label{result with end2end}
\begin{tabular}{ccccc} \\ \hline
\textbf{Top k} &\textbf{BLEU1} &\textbf{BLEU4} & \textbf{ROUGE-L} & \textbf{BERTScore} \\ \hline
1                         & 58.60               & 50.20                     & 63.34                       & 77.13                         \\
5                         & 66.29                & 56.75                     & 71.80                       & 87.33                         \\
10                        & 67.07                    & 57.40                     &  72.61                       & 88.34                         \\
15                        & 67.42                & 57.60                     & 72.98                       & 88.79                         \\
20            & 67.43                     & 57.60           & 72.98             & 88.79                         \\
25                        & 67.58                           & 57.77                                      & 73.16                       & 88.97                         \\
\textbf{30}               & \textbf{67.63}            & \textbf{57.83}            &\textbf{73.23}              & \textbf{89.05}                \\ \hline
\end{tabular}
\end{table}

Based on the results in Section \ref{retriever_result_section}, Section \ref{reader_result_section}, and Section \ref{generator_result_section}, we used a weighted combination of Bi-Encoder and BM25 for the Retriever module, the XLM-R-Large model for the Reader module, and the mBART model for the Generator module to evaluate the performance of the R2GQA system.

Table \ref{result with end2end} demonstrates that as the number of retrieved contexts (top\_k) increases, the performance of system improves. However, the difference between top\_k = 10 and top\_k < 15 is greater than that for top\_k > 10. The performance of system  at top\_k = 10 surpasses that at top\_k = 5 on BLEU1, BLEU4, ROUGE-L, and BERTScore by 0.78\%, 0.65\%, 0.81\%, and 1.01\%, respectively. Similarly, the performance of system at top\_k = 30 exceeds that at top\_k = 10 on BLEU1, BLEU4, ROUGE-L, and BERTScore by 0.56\%, 0.43\%, 0.62\%, and 0.71\%, respectively. Consequently, it can be concluded that the performance nearly reaches saturation at the value of top\_k = 10.

\section{Discussion} \label{discussion}
\subsection{Impact of Context Length In The Reader Module}
To validate the challenge posed by large context length in the dataset, we conducted an experiment to assess the impact of context length on the performance of models in the Reader module. The specific length ranges used for this experiment are detailed in Table \ref{tab:article_length}.

\begin{figure}[!ht]
\centering
\caption{The impact of context length on the performance of models in the Reader module.} \label{fig:context_length_reader}
\includegraphics[width=0.75\textwidth]{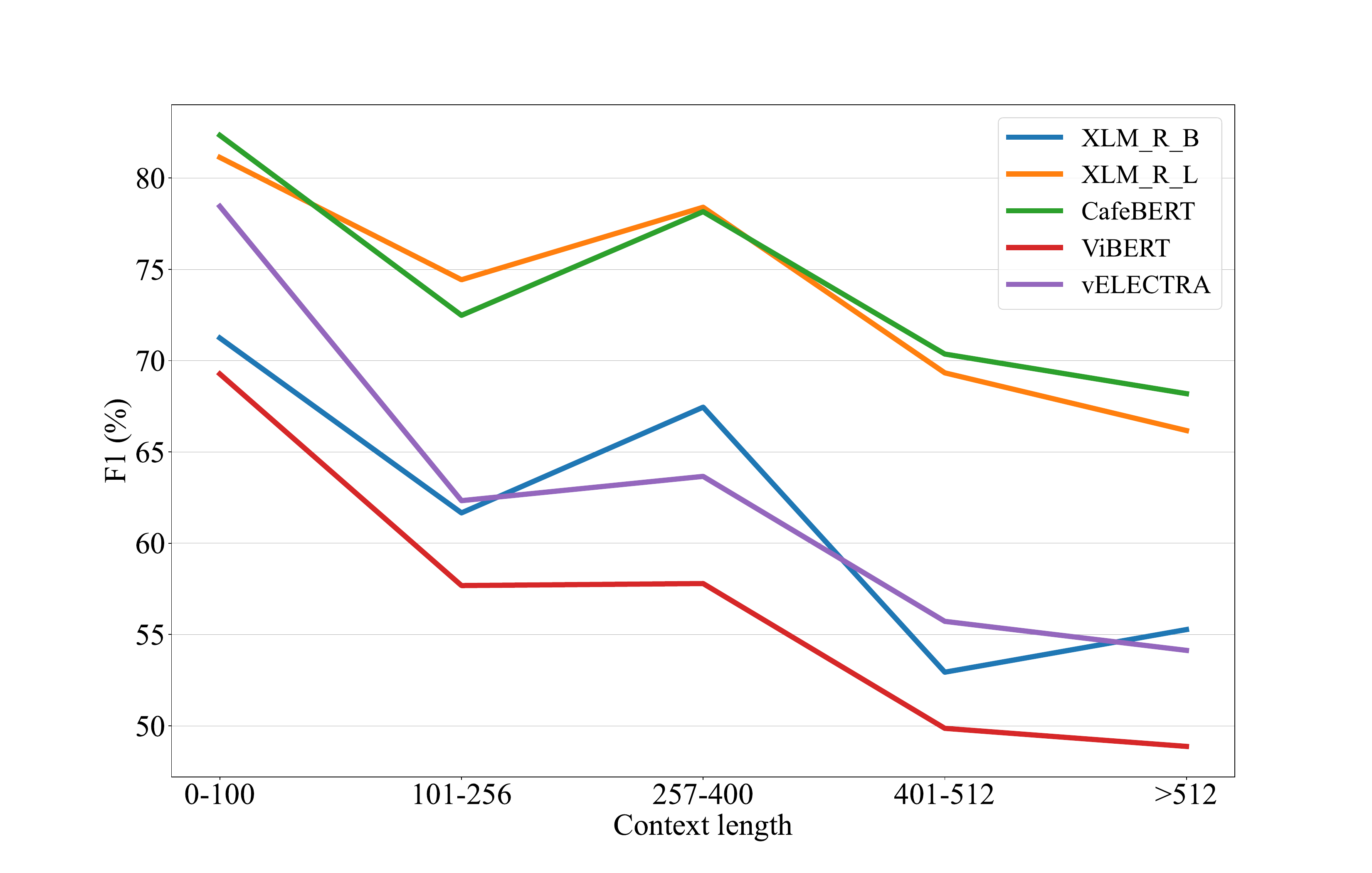}

\end{figure}

Figure \ref{fig:context_length_reader} shows that longer context passages result in lower performance compared to shorter ones. The highest results for all models are achieved in the 0-100 word interval, while the other length intervals yield significantly lower results. In particular, in an interval longer than 512 words, almost all models perform the worst. This length interval exceeds the length limits of all models, causing the models to perform poorly due to the lack of information and context.
\subsection{Impact of Training Sample Number}

To assess the impact of the number of samples in the training set, we trained the model with different quantities: 2000, 4000, 6000, and 7806 questions.

\subsubsection{Machine Reader}
\begin{figure}[!ht]
    \centering
    \includegraphics[width=0.75\textwidth]{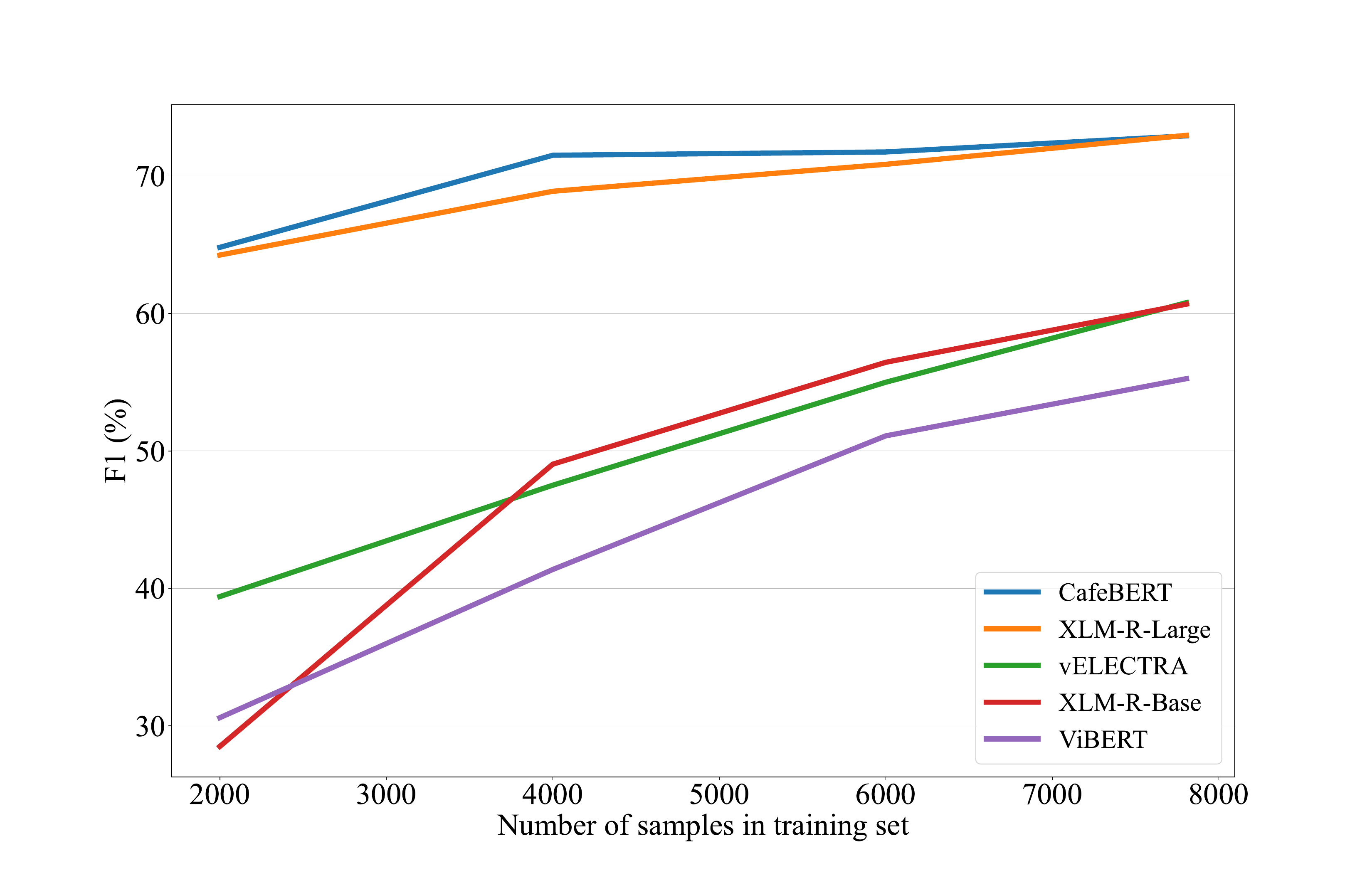}
    \caption{The impact of the amount of training data on the training set of ViRHE4QA on QA models of Reader module.}
    \label{fig:sample_train_reader}
\end{figure}

From Figure \ref{fig:sample_train_reader}, we can observe that increasing the number of samples in the training set significantly improves the performance of all models. Models such as ViBERT and XLM-R-Base show the most significant improvement, whereas models such as CafeBERT and XLM-R-Large show less improvement, as they already perform well even with a small number of training samples. Therefore, the amount of training data has a significant impact on model performance and will continue to increase as the amount of training data increases. Increasing the training data leads to an increase in the number of contexts and vocabulary, which helps the model learn more real-world scenarios.
\subsubsection{Answer Generator}

\begin{figure}[!ht]
    \centering
    \includegraphics[width=0.75\textwidth]{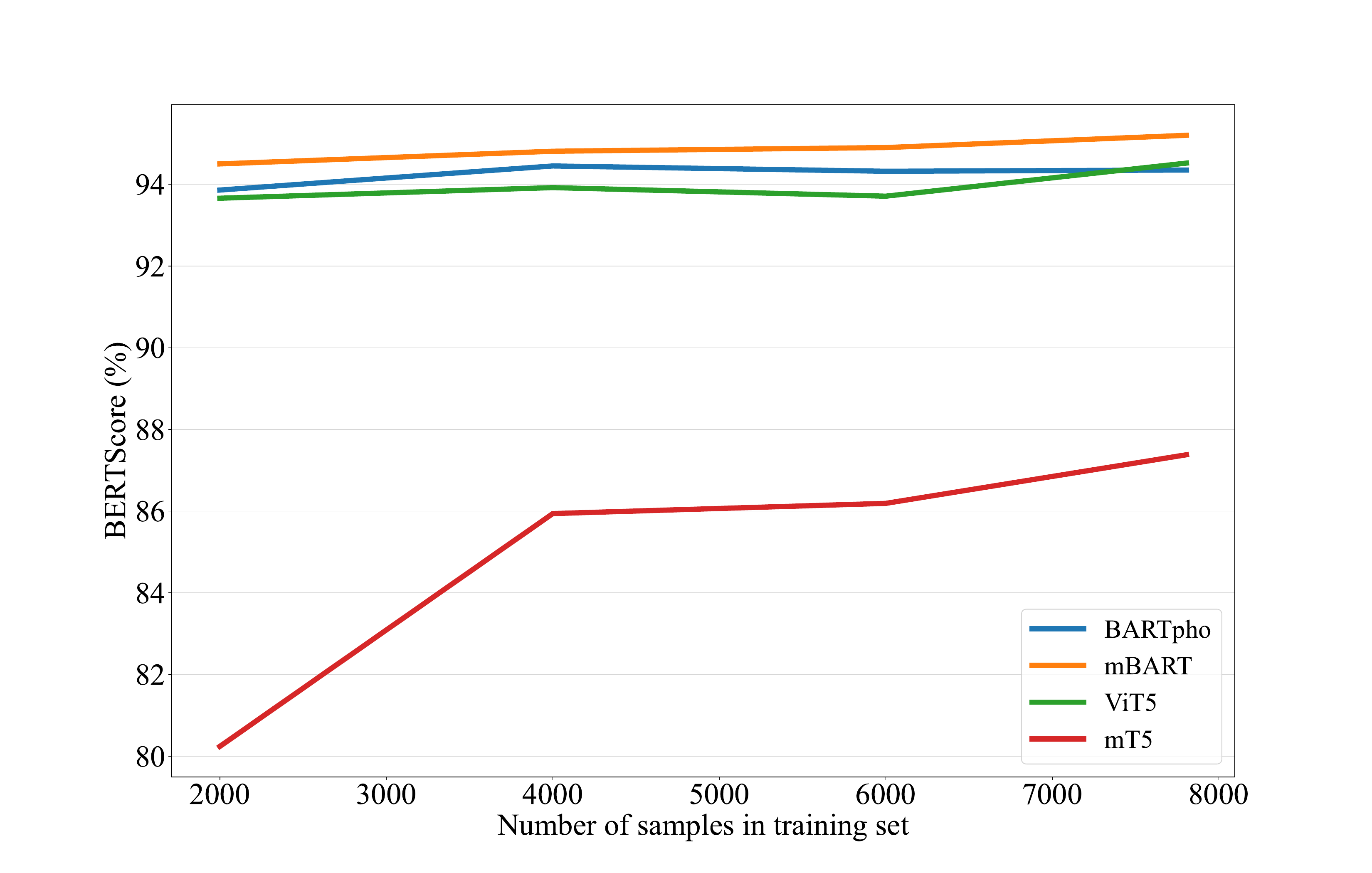}
    \caption{The impact of the amount of training data on the training set of ViRHE4QA on generator models of Generator module.}
    \label{fig:numofgenerator}
\end{figure}

For the Generator module, we focus on the BERTScore metric for analysis. According to Figure \ref{fig:numofgenerator}, it can be observed that as the number of samples in the training set increases, the performance of all models improves slightly. However, for the mT5 model, there is a significant improvement in effectiveness from step 2000 to step 4000; after step 4000, the improvement of the model slows down gradually. This can be explained by the fact that at 2000 data points, the mT5 model has not learned much information yet, but by step 4000, it has accumulated enough information to represent better results. For the BARTpho model, the highest score is achieved when the number of training samples is 4000. It appears that when provided with more data, the information for the model may become noisy, leading to no improvement in results. In contrast to the Reader module, the Generator module shows little improvement with an increasing number of samples in the training set. Performance curves remain relatively flat as the training set size increases. This indicates that the transformer models in this module perform very well with less data.

\subsection{Impact of Context in Answer Generator Module}

We experimented with the influence of context on the Generator module using different types of input: Question and Context (Q+C); Question, Extractive answer, and Context (Q+E+C); Question and Extractive answer (Q+E). We added the token </s> to the model input as described in Section \ref{experiment_generator}.

\begin{figure}
    \centering
    \includegraphics[width=0.75\linewidth]{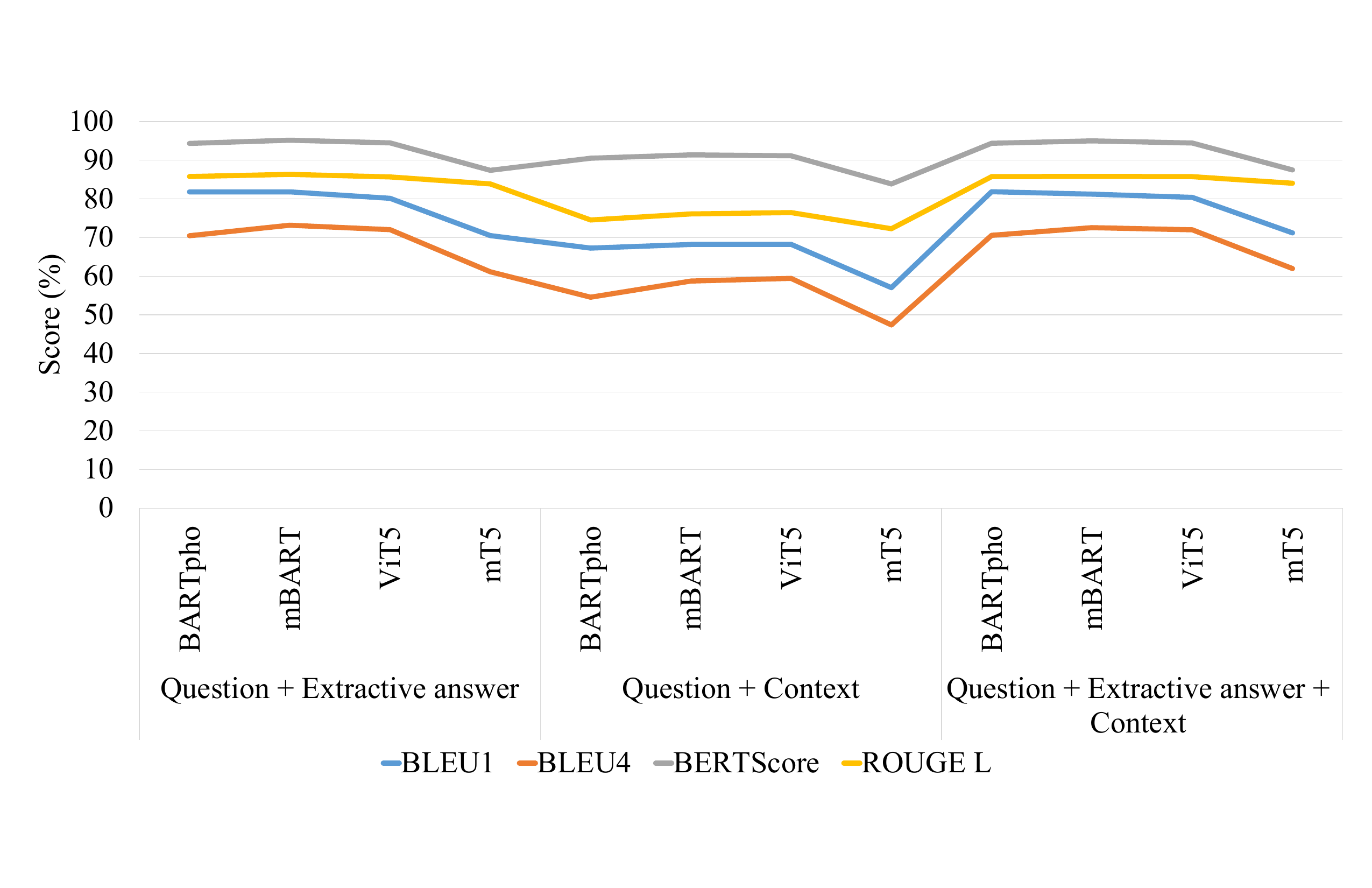}
    \caption{Compare the impact of context on the Generator module.}
    \label{fig:context_in_generator}
\end{figure}

From Table \ref{tab:generator_module}, Table \ref{tab:generator_module_question_context}, Table \ref{tab:generator_module_question_extract_context}, and Figure \ref{fig:context_in_generator}, it can be observed that the performance of the Generator module is highest when the input is Q+E, followed by Q+E+C, and lowest for Q+C. The differences between Q+E and Q+E+C are not significant but are markedly higher than the scores for Q+C. This section reveals that incorporating context into the module is not particularly effective and increases the input length to the Generator module, leading to inaccurate results.

\begin{table}[]
\centering
\caption{Results of the Generator module with input are a Question and a Context.}
\label{tab:generator_module_question_context}
\begin{tabular}{lcccc}
\hline
\textbf{Model} & \textbf{BLEU1} & \textbf{BLEU4} & \textbf{BERTScore} & \textbf{ROUGE-L} \\ \hline
BARTpho & 67.28 & 54.58 & 90.53 & 74.54 \\
mBART & 68.23 & \textbf{58.76} & \textbf{91.42} & 76.11 \\
ViT5 & \textbf{68.25} & 59.45 & 91.14 & \textbf{76.48} \\
mT5 & 57.05 & 47.40 & 83.87 & 72.29 \\ \hline
\end{tabular}
\end{table}

\begin{table}[]
\centering
\caption{Results of the Generator module with input are a Question, an Extractive answer and a Context.}
\label{tab:generator_module_question_extract_context}
\begin{tabular}{lcccc}
\hline
\textbf{Model} & \textbf{BLEU1} & \textbf{BLEU4} & \textbf{BERTScore} & \textbf{ROUGE-L} \\ \hline
BARTpho & \textbf{81.88} & 70.57 & 94.39 & 85.76 \\
mBART & 81.24 & \textbf{72.58} & \textbf{95.03} & \textbf{85.82} \\
ViT5 & 80.38 & 72.01 & 94.47 & 85.76 \\
mT5 & 71.23 & 61.98 & 87.53 & 84.07 \\ \hline
\end{tabular}
\end{table}

For the Q+E method, the input is more concise, focusing on the main point (extractive answer) to provide the final answer for the system. In addition, using a shorter input reduces costs and resources when operating the system.

\subsection{Performance of QA Systems}

In this section, we will compare the performance of our system with other QA systems, Naive RAG. The main metrics used for the evaluation include BLEU1, BLEU4, ROUGE1, ROUGE-L, and BERTScore. The experiments were conducted on an NVIDIA RTX 3090 GPU with 24GB VRAM from VastAI\footnote{\url{https://vast.ai/}}.

\begin{figure}
    \centering
    \includegraphics[width=0.75\linewidth]{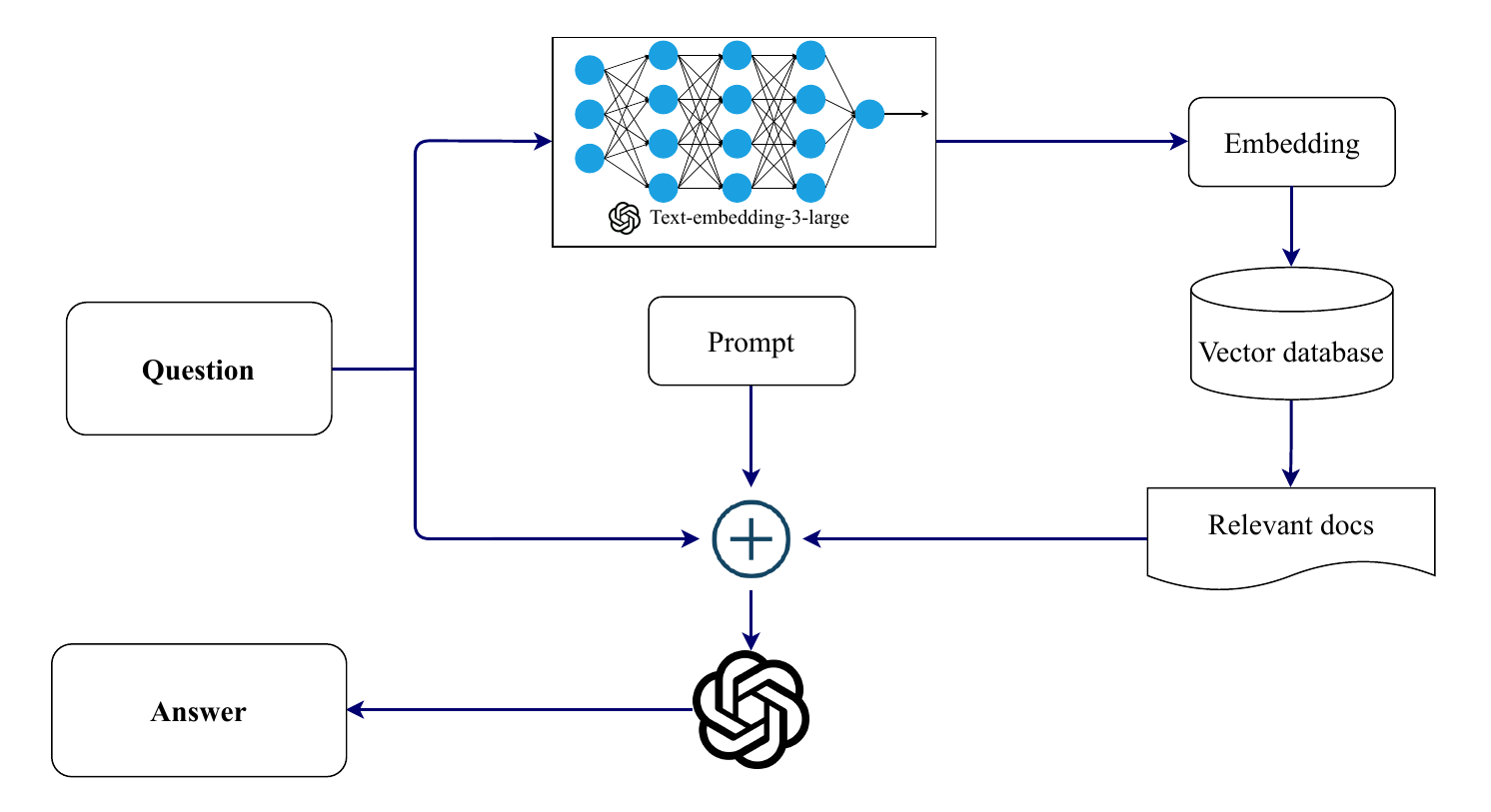}
    \caption{Flowchart of a Retrieval-Augmented Generation (RAG) system. }
    \label{fig:ragsystem}
\end{figure}

\textbf{System Configurations:}
\begin{quote}
    \textbf{Naive RAG:} RAG is a question-answering system proposed by \cite{RAG}. In the past two years, RAG has been widely used as large language models (LLMs) have developed. Therefore, we compare our system with this method. We use the $text-embedding-3-large$ model to create the vector database and encode the questions. The vector database we use is Chromadb, designed by Langchain, and the large language model used to generate the final answer is $GPT-3.5-turbo-instruct$. For the prompt method, we use one-shot prompting, meaning that the prompt includes one sample example and the question to be answered. The model can refer to the sample example to answer the question more accurately. We use the $text-embedding-3-large$ model and the $GPT-3.5-turbo-instruct$ model from Microsoft Azure AI\footnote{\url{https://azure.microsoft.com/en-us/solutions/ai}}. The system is illustrated in Figure \ref{fig:ragsystem}.
\end{quote}

\textbf{Comparison Results:}

Table \ref{tab:PerformanceOfQASys}  shows that our system achieves higher performance compared to Naive RAG across most metrics, specifically with only top\_k = 1. In addition to comparing response times, we also consider operational costs. For the RAG system, the use of APIs incurs \$0.021 for the vectorization of the database and \$0.0029 per question. In contrast, our system does not generate operational costs per answer. This indicates the high potential application of the R2GQA in real-world QAs tasks.
\begin{table}[!ht]
\caption{Results of two systems on ours test set.} \label{tab:PerformanceOfQASys}
\resizebox{1\linewidth}{!}{
\begin{tabular}{lrcccccc} \\ \hline
           & \textbf{Top k} & \textbf{BLEU1 (\%)} & \textbf{BLEU4 (\%)} & \textbf{ROUGE-L (\%)} & \textbf{BERTScore (\%)}      & \textbf{Cost (\$/question)} \\ \hline
  R2QGA   &  1                        & 56.19                      & 47.09                     & 59.85                       & 78.82                                &  - \\
Naive-RAG                     & 1                        & 51.18   & 38.27  & 59.75   & 84.43         &  0.0029  \\ \hline
\end{tabular}}
\end{table}
\section{Error Analysis}\label{sec:error_analysis}

To perform error analysis, we surveyed 200 question-answer pairs predicted by the R2GQA system from the test set and classified errors into the following 5 main types:

\textbf{Repetition in extractive answers:} in the Reader module, we use a BIO format model, which often leads to extractive answers containing one or more repeated words, resulting in grammatically incorrect phrases in Vietnamese. 
\begin{quote}

\begin{quote}
\textbf{Context:} \\
Điều 1. Phạm vi điều chỉnh và đối tượng áp dụng\\
1. Mỗi học kỳ, Trường Đại học A (Trường) tổ chức \textbf{01} lần thi lý thuyết giữa kỳ và \textbf{01 lần} thi lý thuyết cuối kỳ tập trung trực tiếp hoặc trực tuyến (gọi chung là thi) cho các môn học được mở trong học kỳ đó. Thời gian tổ chức thi được qui định trên biểu đồ giảng dạy năm học. Quy định này quy định chung về việc tổ chức các đợt thi bao gồm các công tác: chuẩn bị thi, tổ chức thi, chấm thi, công bố điểm, phúc khảo, chế độ lưu trữ và xử lý vi phạm. Quy định này áp dụng đối với hệ đại học chính quy. Các chương trình đặc biệt có thể có kế hoạch thi riêng tùy theo đặc thù của chương trình.\\
2. Các hình thức thi bao gồm:\\
Tự luận (hoặc tự luận kết hợp với trắc nghiệm) \\
Trắc nghiệm \\
Vấn đáp \\
Đồ án\\
3. Việc tổ chức thi theo hình thức trực tuyến thực hiện theo quy định riêng. \\
\textit{(Article 1. Scope of Regulation and Applicable Subjects\\
1. Each semester, the University A (the University) organizes one mid-term theoretical exam and one final theoretical exam, either in-person or online (collectively referred to as exams), for the courses offered in that semester. The exam schedule is determined on the academic year teaching schedule. This regulation provides general provisions on organizing exam sessions, including exam preparation, administration, grading, result announcement, appeals, storage, and violation handling. This regulation applies to regular undergraduate programs. Special programs may have their own exam plans depending on the program's characteristics.\\
2. Exam formats include:\\
\textit{Essay (or a combination of essay and multiple-choice)}\\
\textit{Multiple-choice}\\
\textit{Question and answer}\\
\textit{Project}\\
3. Online exam organization follows separate regulations.)}\
\\
\textbf{Question:} 
    Mỗi học kỳ, Trường Đại A (Trường) tổ chức mấy lần thi lý thuyết cuối kỳ cho các môn học được mở trong học kỳ đó? \textit{(How many end-of-term theoretical exams does the University A organize for subjects offered in each semester?)}\
\\
\textbf{Predicted extractive:} 
    01 01 lần \textit{(01 01 times)}
\\
\textbf{True extractive:}
    01 \textit{(01)}
\end{quote}
\end{quote}

\textbf{Incorrect information extraction:} this error occurs when the Reader module extracts inaccurate information or information that does not match the context of the question.\
\begin{quote}
\begin{quote}
\textbf{Context:} \\
    Điều 14. Chế độ lưu trữ\\
    Toàn bộ biên bản, hồ sơ bảo vệ KLTN được Thư ký Hội đồng bàn giao cho Khoa và được các Khoa lưu trữ tối thiểu trong vòng năm năm. \\
    Khoa tổng hợp điểm vào danh sách các SV đã đăng ký làm KLTN (kể cả các SV không hoàn thành KLTN) do P. ĐTĐH cung cấp và gởi lại cho P. ĐTĐH \textbf{không quá 2 tuần sau ngày bảo vệ.}\\ 
    \textit{(Article 14. Storage Regime\\
    The entire minutes, records of thesis defense are handed over by the Council Secretary to the Faculty and are archived by the Faculties for a minimum of five years.\\
    The Faculty consolidates the scores into the list of students who have registered to do their thesis (including students who have not completed their thesis) provided by the Academic Affairs Office and returns it to the Academic Affairs Office no later than 2 weeks after the defense date.)}\

\textbf{Question:} Thời gian để Khoa tổng hợp điểm vào danh sách sinh viên tham gia KLTN và gởi cho P. ĐTĐH được tính từ khi nào? \textit{(From when is the time for the Faculty to compile grades into the list of students participating in the graduation thesis and send it to the Undergraduate Training Department calculated?)}\
\\
\textbf{Predicted extractive:} không quá 2 tuần sau ngày bảo vệ. \textit{(not more than 2 weeks after the defense day.)}\
\\
\textbf{True extractive:} sau ngày bảo vệ. \textit{(after the defense day.)}\
\\
\textbf{Predicted abstractive:} Thời gian để Khoa tổng hợp điểm vào danh sách sinh viên tham gia KLTN và gởi cho P. ĐTĐH không quá 2 tuần sau ngày bảo vệ. \textit{(The time for the Faculty to compile grades into the list of students participating in the graduation thesis and send it to the Undergraduate Training Department is not more than 2 weeks after the defense day.)}\
\\
\textbf{True abstractive:} Thời gian để Khoa tổng hợp điểm vào danh sách sinh viên tham gia KLTN và gởi cho P. ĐTĐH được bắt đầu tính sau ngày bảo vệ. \textit{(The time for the Faculty to compile grades into the list of students participating in the graduation thesis and send it to the Undergraduate Training Department is calculated from the day after the defense.)}
\end{quote}
\end{quote}

\textbf{Over-extraction/Under-extraction of information:} this error occurs when the system extracts more or less information than required by the question, confusing for the user.\
\begin{quote}
\begin{quote}
\textbf{Context:} \\
    Điều 17. Quản lý điểm thi\\
    1.	\textbf{Cán bộ chấm thi} chịu \textbf{trách} nhiệm về tính chính xác của thông tin điểm thi được nhập và công bố cho SV trên Hệ thống quản lý điểm trong thời hạn chấm thi và nộp điểm.\\
    2.	P.ĐTĐH/VPĐB chịu trách nhiệm kiểm tra thông tin điểm thi trên Hệ thống quản lý điểm so với điểm được ghi trên bài thi, tiếp nhận và xử lý khiếu nại của SV về điểm thi và cấp bảng điểm theo yêu cầu.\\
    3.	Phòng Dữ liệu và Công nghệ Thông tin có trách nhiệm đảm bảo an toàn cho dữ liệu điểm trên Hệ thống quản lý điểm; đảm bảo chỉ cấp quyền nhập và chỉnh sửa điểm cho cán bộ chấm thi đối với thành phần điểm và lớp mà mình phụ trách chấm thi trong thời hạn chấm thi và nộp điểm. Mọi thao tác trên dữ liệu điểm phải được ghi nhận lại đầy đủ và chính xác (người nhập điểm, người chỉnh sửa điểm, thời gian và lý do chỉnh sửa). \\
    \textit{(Article 17. Examination Score Management\\
    1. The examiners are responsible for the accuracy of the exam score information entered and announced to students on the Grade Management System within the grading and score submission period.\\
    2. The Academic Affairs Office/Department is responsible for verifying the exam score information on the Grade Management System against the scores recorded on the exam papers, receiving and handling student complaints about exam scores, and issuing transcripts upon request.\\
    3. The Data and Information Technology Department is responsible for ensuring the security of the exam score data on the Grade Management System, ensuring that only examiners responsible for grading their assigned components and classes have the authority to enter and edit scores within the grading and score submission period. All operations on exam score data must be recorded accurately and completely (person entering the score, person editing the score, time, and reason for the edit).)}\\

\textbf{Question:} Phòng Dữ liệu và Công nghệ Thông tin cấp quyền nhập và chỉnh sửa điểm cho ai? \textit{(To whom does the Data and Information Technology Department grant the right to enter and edit scores?)}\

\textbf{Predicted extractive:} cán bộ chấm thi trách \textit{(exam graders responsible)}\

\textbf{True extractive:} cán bộ chấm thi \textit{(exam graders)}\

\textbf{Predicted abstractive:} Phòng Dữ liệu và Công nghệ Thông tin cấp quyền nhập và chỉnh sửa điểm cho cán bộ chấm thi trách. \textit{(The Data and Information Technology Department grants the right to enter and edit scores to the responsible exam graders.)}\

\textbf{True abstractive:} Phòng Dữ liệu và Công nghệ Thông tin chỉ cấp quyền nhập và sửa điểm cho cán bộ chấm thi. \textit{(The Data and Information Technology Department grants the right to enter and edit scores only to the exam graders.)}
\end{quote}
\end{quote}

\textbf{Incorrect context extraction:} the retrieval system returns a text segment that is unrelated or does not contain the necessary information to answer the question.\
\begin{quote}
\begin{quote}
\textbf{Question:} BĐH tổ chức lấy ý kiến sinh viên về việc gì? \textit{(What does the Management Board organize to collect student opinions about?)}\

\textbf{True context:} \\
    Điều 10. Giảng dạy các môn CTTN\\
    CTTN phải được thực hiện trên quan điểm lấy người học làm trung tâm. Người học phải được tạo điều kiện để thể hiện vai trò chủ động trong tiến trình học tập. Người học phải đóng vai trò chủ động trong hoạt động học tập, thay vì thụ động tiếp nhận kiến thức.\\
    Sinh viên CTTN sẽ học cùng với sinh viên các lớp chương trình chuẩn trong các môn được đào tạo chung, các môn học cốt lõi dành riêng cho sinh viên CTTN được tổ chức lớp học riêng.\\
    Khoa quản lý chuyên môn có trách nhiệm chọn các cán bộ có kinh nghiệm để phụ trách giảng dạy. Các môn học tài năng và KLTN phải do CBGD có học vị tiến sĩ hoặc giảng viên chính, hoặc thạc sĩ tốt nghiệp ở các trường Đại học thuộc các nước tiên tiến, đúng ngành hoặc thuộc ngành gần đảm nhiệm.\\
    Trong tuần đầu tiên của học kỳ, CBGD phải thông báo công khai cho sinh viên về đề cương giảng dạy môn học; trong đó đặc biệt chú ý các thông tin, các phần học bổ sung tăng cường; số cột điểm và tỷ lệ tính của từng cột điểm vào điểm tổng kết môn học.\\
    CBGD phải cung cấp đầy đủ đề cương môn học, tài liệu và công bố nội dung bài giảng trước cho sinh viên trên trang web môn học. 	\\
    Đầu mỗi học kỳ, đại diện đơn vị quản lý chương trình và các CVHT phải gặp gỡ đại diện sinh viên (ít nhất 3 SV/lớp – do lớp bầu chọn) tất cả các lớp CTTN để trao đổi và nhận phản hồi về tình hình giảng dạy và sinh hoạt. Cuối học kỳ, BĐH phối hợp với phòng Thanh tra - Pháp chế - Đảm bảo chất lượng tổ chức lấy ý kiến sinh viên (dùng phiếu thăm dò, qua trang web,…) về giảng dạy môn học và tổ chức cho CBGD rút kinh nghiệm về các góp ý của sinh viên.\\
    Ngoài nội dung bắt buộc theo đề cương, các môn CTTN có thể có thêm các nội dung tăng cường và một số lượng hạn chế các buổi ""seminar ngoại khóa"". Lịch dạy và lịch dạy bổ sung tăng cường, dạy bù được báo cáo và kiểm tra theo quy trình chung như lớp đại học chính quy đại trà. \\
    \textit{(Article 10. Teaching CTTN Courses\\
    Teaching CTTN must be based on the learner-centered approach. Learners must have conditions to play an active role in the learning process. Learners must take an active role in learning activities, rather than passively receiving knowledge.\\
    CTTN students will study alongside students of standard programs in jointly taught subjects; core subjects specifically for CTTN students will have separate class arrangements.\\
    The specialized management department is responsible for selecting experienced personnel to teach. Talent courses and thesis projects must be taught by instructors with a doctoral degree or lecturers who have graduated from universities in advanced countries, in the relevant field or closely related fields.\\
    In the first week of the semester, instructors must publicly announce to students the course syllabus, paying particular attention to additional information, supplementary learning sections, the number of columns for grading, and the weighting of each column in the overall grade of the course.\\
    Instructors must provide complete course syllabi, materials, and pre-announce lecture contents to students on the course website.\\
    At the beginning of each semester, program management representatives and class advisors must meet with student representatives (at least 3 students per class – elected by the class) from all CTTN classes to exchange and receive feedback on teaching and activities. At the end of the semester, the Faculty Board must coordinate with the Inspection and Quality Assurance Department to collect student opinions (using surveys, websites, etc.) on course teaching and organize instructors to learn from student feedback.\\
    In addition to the mandatory content outlined in the syllabus, CTTN courses may include additional supplementary content and a limited number of extracurricular seminars. Teaching schedules and additional teaching schedules, makeup classes must be reported and monitored according to the general procedures for regular undergraduate classes.)}

\textbf{Predicted context:} \\
    Điều 16. Đảm bảo chất lượng\\
    - Đơn vị chuyên môn có trách nhiệm chọn các cán bộ đạt yêu cầu theo quy định và có kinh nghiệm giảng dạy để phụ trách giảng dạy các môn học cho các lớp thuộc CT CLC.\\
    - Trong vòng 02 tuần đầu của mỗi học kỳ, đơn vị quản lý và các cố vấn học tập gặp gỡ sinh viên tất cả các lớp CT CLC để trao đổi và nhận phản hồi về tình hình giảng dạy và sinh hoạt.\\
    - Cuối học kỳ, đơn vị quản lý phối hợp với phòng Thanh tra - Pháp chế - Đảm bảo chất lượng tổ chức lấy ý kiến sinh viên (dùng phiếu thăm dò, qua website, …) về việc giảng dạy môn học và có trách nhiệm trao đổi với giảng viên để rút kinh nghiệm về các góp ý của SV.\\
    - Tổ chức lấy ý kiến sinh viên tối thiểu 01 lần mỗi học kỳ về \textbf{công tác quản lý, phục vụ, điều kiện cơ sở vật chất và} tổ chức trả lời những ý kiến phản ánh của sinh viên.\\ 
    \textit{(Article 16. Quality Assurance\\
    - The specialized unit is responsible for selecting qualified personnel as required by regulations and with teaching experience to be in charge of teaching courses for classes under the CT CLC program.\\
    - Within the first two weeks of each semester, the management unit and academic advisors meet with students from all CT CLC classes to exchange and receive feedback on teaching and activities.\\
    - At the end of the semester, the management unit collaborates with the Inspection and Quality Assurance Department to collect student opinions (via surveys, websites, etc.) on course teaching and is responsible for exchanging feedback with instructors to learn from student suggestions.\\
    - Student feedback sessions should be organized at least once per semester on management activities, services, facilities, and addressing student feedback.)}\
    
\textbf{Predicted extractive:} công tác quản lý, phục vụ, điều kiện cơ sở vật chất và \textit{(management work, services, facilities, and)}\

\textbf{True extractive:} giảng dạy môn học \textit{(teaching of the course)}\

\textbf{Predicted abstractive:} BĐH tổ chức lấy ý kiến sinh viên về công tác quản lý, phục vụ, điều kiện cơ sở vật chất và các vấn đề khác có liên quan đến việc dạy bù từ cán bộ giảng dạy. \textit{(The Management Board organizes to collect student opinions about management work, services, facilities, and other issues related to makeup teaching by lecturers.)}\

\textbf{True abstractive:} BĐH tổ chức lấy ý kiến sinh viên về giảng dạy môn học. \textit{(The Management Board organizes to collect student opinions about the teaching of the course.)}\
\end{quote}
\end{quote}

\textbf{No information in the Machine Reader module:} The approach of the Reader Module is sequence tagging, hence there are instances where information cannot be extracted from the retrieved context.

The R2GQA system performs well in generating user-friendly free-form answers, but it still contains many basic errors. These errors include repeating information, extracting redundant or missing information, retrieving incorrect context segments, and extracting information from inaccurate context segments. These issues reduce the accuracy and efficiency of the system, potentially causing confusion for users.

\section{Conclusion and Future Work}  \label{sec:conclusion_future_work}

In this paper, we introduce the ViRHE4QA dataset, a QA dataset based on academic regulations in higher education. The dataset comprises 9,758 meticulously constructed data samples, created by seven well-trained and closely monitored annotators. Furthermore, we proposed the R2GQA system, which consists of three modules: Retrieval, Reader, and Generator. This system leverages state-of-the-art language models, delivering high performance and efficiency for the QA task without relying on any third-party services. We conducted various experiments on both the dataset and the system to demonstrate their effectiveness and practical applicability.

However, the ViRHE4QA dataset and the R2GQA system face several challenges that need to be addressed in the future. These challenges include managing the length of the input data and improving the accuracy of the Retrieval and Reader systems. We also plan to focus on optimizing and fine-tuning the language models to better fit the context and linguistic characteristics of the Vietnamese language.

Moreover, expanding and enriching the dataset with a broader variety of questions and diverse contexts will enhance the versatility of the system. We aim to explore and integrate advanced techniques such as deep learning and reinforcement learning to further improve the accuracy and performance of the system. Reducing processing time while maintaining high accuracy is a crucial goal in ensuring the practical application of automated QA systems.

Finally, deploying the system in real-world environments will provide valuable feedback, helping us to understand its strengths and weaknesses and propose appropriate improvements. We hope that this work will make a significant contribution to the field of automated QA and open up new research directions in the future.

\section*{Acknowledgement}
This research was supported by The VNUHCM-University of Information Technology's Scientific Research Support Fund.

\section*{Declarations}

\textbf{Conflict of interest} The authors declare that they have no conflict of interest.

\section*{Data Availability}

Data will be made available on reasonable request.

\bibliography{sn-bibliography}

\end{document}